\documentclass[10pt,journal,compsoc]{IEEEtran}

\usepackage{cite}

\ifCLASSINFOpdf

\else

\fi

\usepackage{times}  
\usepackage{helvet} 
\usepackage{courier}  
\usepackage[hyphens]{url}  
\usepackage{graphicx} 
\usepackage{graphicx}  
\usepackage{amsmath,graphicx}

\usepackage{amsmath}
\usepackage{color,xcolor,soul,framed} 
\usepackage{xspace}
\usepackage{array}
\usepackage{eqparbox}
\usepackage{url}
\usepackage{longtable}
\usepackage{lipsum}
\usepackage{blindtext}
\usepackage{makecell}
\usepackage{mathtools}
\usepackage{commath}
\usepackage{booktabs}
\usepackage{multirow}
\usepackage{tabularx}
\usepackage[normalem]{ulem}
\usepackage{xspace}
\usepackage{algorithm}
\usepackage{algpseudocode}
\usepackage{amsfonts}
\usepackage{placeins}

\usepackage[normalem]{ulem}
\usepackage{pythonhighlight}

\newcolumntype{C}{>{\hsize=\dimexpr0.5\hsize+8\tabcolsep+\arrayrulewidth\centering\relax}X}

\DeclareMathOperator*{\argmax}{arg\,max}

\DeclareMathOperator{\cref}{ref}

\DeclareMathOperator{\softmax}{softmax}

\DeclarePairedDelimiter\floor{\lfloor}{\rfloor} 
\DeclarePairedDelimiter{\nint}\lfloor\rceil
\newcommand{\convMatrix}[3]{
\begin{math}
\begin{bmatrix}
\text{Conv1D},\text{kernel}(#1), \text{output channels}(#2) \\
\text{BatchNorm1D},\text{output channels}(#3) \\ 
\text{LeakyReLU},\text{slope(0.3)}\ 
\end{bmatrix} 
\end{math}
}

\newcommand{\convMatrixc}[1]{
\begin{math}
\begin{bmatrix}
\text{Linear}, 64 \\
\text{ReLU}, - \\ 
\text{Dropout},0.5\\ 
\text{Linear},#1\ 
\end{bmatrix} 
\end{math}
}

\usepackage{listings}
\usepackage[frozencache=true,cachedir=minted-cache]{minted} 

\usepackage[pagebackref=true,breaklinks=true,colorlinks,bookmarks=false]{hyperref}
\hypersetup{
    colorlinks=true,
    filecolor=magenta,      
    urlcolor=cyan,
}

\begin{document}

\title{PARSE: Pairwise Alignment of Representations in Semi-Supervised EEG Learning for\\Emotion Recognition}

\author{Guangyi~Zhang, Vandad~Davoodnia,  
        and~Ali~Etemad,~\IEEEmembership{Senior Member,~IEEE}
\thanks{G. Zhang, V. Davoodnia, and A. Etemad are with the Department of Electrical and Computer Engineering, and Ingenuity Labs Research Institute, Queen's University, Kingston, ON, Canada (e-mail: \{guangyi.zhang, vandad.davoodnia, ali.etemad\}@queensu.ca).}

}

\maketitle

\bstctlcite{IEEEexample:BSTcontrol}

\begin{abstract}
We propose PARSE, a novel semi-supervised architecture for learning reliable EEG representations for emotion recognition. To reduce the potential distribution mismatch between large amounts of unlabeled data and a limited number of labeled data, PARSE uses pairwise representation alignment. First, our model performs data augmentation followed by label guessing for large amounts of original and augmented unlabeled data. The model is then followed by sharpening the guessed labels and convex combinations of the unlabeled and labeled data. Finally, it performs representation alignment and emotion classification. To rigorously test our model, we compare PARSE to several state-of-the-art semi-supervised approaches, which we implement and adapt for EEG learning. We perform these experiments on four public EEG-based emotion recognition datasets, SEED, SEED-IV, SEED-V and AMIGOS (valence and arousal). The experiments show that our proposed framework achieves the overall best results with varying amounts of limited labeled samples in SEED, SEED-IV and AMIGOS (valence), while approaching the overall best result (reaching the second-best) in SEED-V and AMIGOS (arousal). The analysis shows that our pairwise representation alignment considerably improves the performance by performing the distribution alignment between unlabeled and labeled data, especially when only $1$ sample per class is labeled. The source code of our paper is publicly available at
\href{https://github.com/guangyizhangbci/PARSE}{https://github.com/guangyizhangbci/PARSE}.
\end{abstract}


\section{Introduction} \label{sec:intro}
Human emotions are encountered and experienced by humans on a daily basis, and significantly influence our behaviors, interactions, and perceptions of the world. It is, therefore, critical to develop algorithms capable of recognizing different human emotions from behavioural measures and physiological signals to assist computers in better responding to human needs and interactions. 
As a result, affective computing, a discipline that aims to develop data-driven computational models capable of recognizing human emotional states \cite{picard2000affective}, has become a popular field of research in recent years.

Emotions can be detected and quantified using a variety of different human-generated signals, including facial expressions \cite{sepas2020facial}, speech \cite{hajavi2020knowing}, and bio-signals such as electrocardiogram (ECG) \cite{sarkar2020self}, electrodermal activity\cite{bhatti2021attentive}, photoplethysmogram \cite{hassan2019human}, Electroencephalogram (EEG) \cite{zhang2021distilling}, and others. While each modality comes with a unique set of advantages and disadvantages (e.g., wearability, cost, accuracy, etc.), EEG is highly regarded as an informative and viable option for affective computing given its direct relation to the central nervous system.

Deep learning models have become increasingly popular in recent years due to the significant advancements in deep neural networks, computational power, and abundance of collected data. Such methods have shown great promise in dealing with various challenges encountered in EEG, including high dimensionality, non-stationarity, and high susceptibility to noise and the linear mixing effect \cite{zhang2019classification,zhang2020rfnet}. However, majority of these methods are fully-supervised and rely heavily on large amounts of `labeled' training samples.
On the other hand, labeling EEG signals is challenging, time-consuming, expensive, and often requires annotation experts. For example, many EEG-based works in the area require diverse emotion annotating methods such as pre-stimulation self-assessment, post-experiment self-assessment, and numerous expert evaluations to acquire accurate labels \cite{zheng2015investigating,zheng2018emotionmeter,liu2021comparing,correa2018amigos}. When only a very small portion of the training samples are labeled, most existing fully-supervised learning solutions suffer from performance degradation. To address this challenge, we propose a novel Semi-Supervised Learning (SSL) pipeline by efficiently leveraging large amounts of unlabeled EEG samples and very few labeled ones.

\noindent \textbf{Problem statement.}
(\textbf{1}) The field of SSL has witnessed interesting progress in recent years, most notably in the computer vision domain \cite{samuli2017temporal,tarvainen2017mean,oliver2018realistic,berthelot2019mixmatch,sohn2020fixmatch,berthelot2021adamatch}. Nonetheless, very few works have adopted these state-of-the-art methods in the field of EEG representation learning, and fewer so have proposed \textit{novel} SSL frameworks to learn EEG for emotion recognition or otherwise. (\textbf{2}) In most recent SSL methods, the distribution of unlabeled samples plays a critical role in the performance of the model. In particular, when pseudo-labels are generated (guessed) for unlabeled samples, existing methods often consider the confidence of the model on the estimated pseudo-labels to accept or reject unlabeled samples for which low-confidence pseudo-labels cannot be guessed \cite{sohn2020fixmatch,berthelot2021adamatch}. While this approach is logically viable and practically effective, the confidence threshold set is often dataset-specific and requires trial and error. In particular, this would be a significant challenge given the distribution differences between different datasets in the context of EEG. (\textbf{3}) In addition to the internal distribution of unlabeled data,
the closeness of the distribution of unlabeled samples with respect to \textit{labeled} ones plays a key role in SSL methods as well. Should the distribution of labeled and unlabeled samples be far from each other, the quality of the generated pseudo-labels for the unlabeled samples might get compromised due to the lack of generalizability in the model. While most SSL methods have not addressed this issue \cite{berthelot2019mixmatch,sohn2020fixmatch,berthelot2021adamatch}, we believe that this problem needs more attention, especially in the context of semi-supervised EEG representation learning, given the vast distribution differences often found within different datasets.

\begin{figure}
    \begin{center}
    \includegraphics[width=1.0\columnwidth]{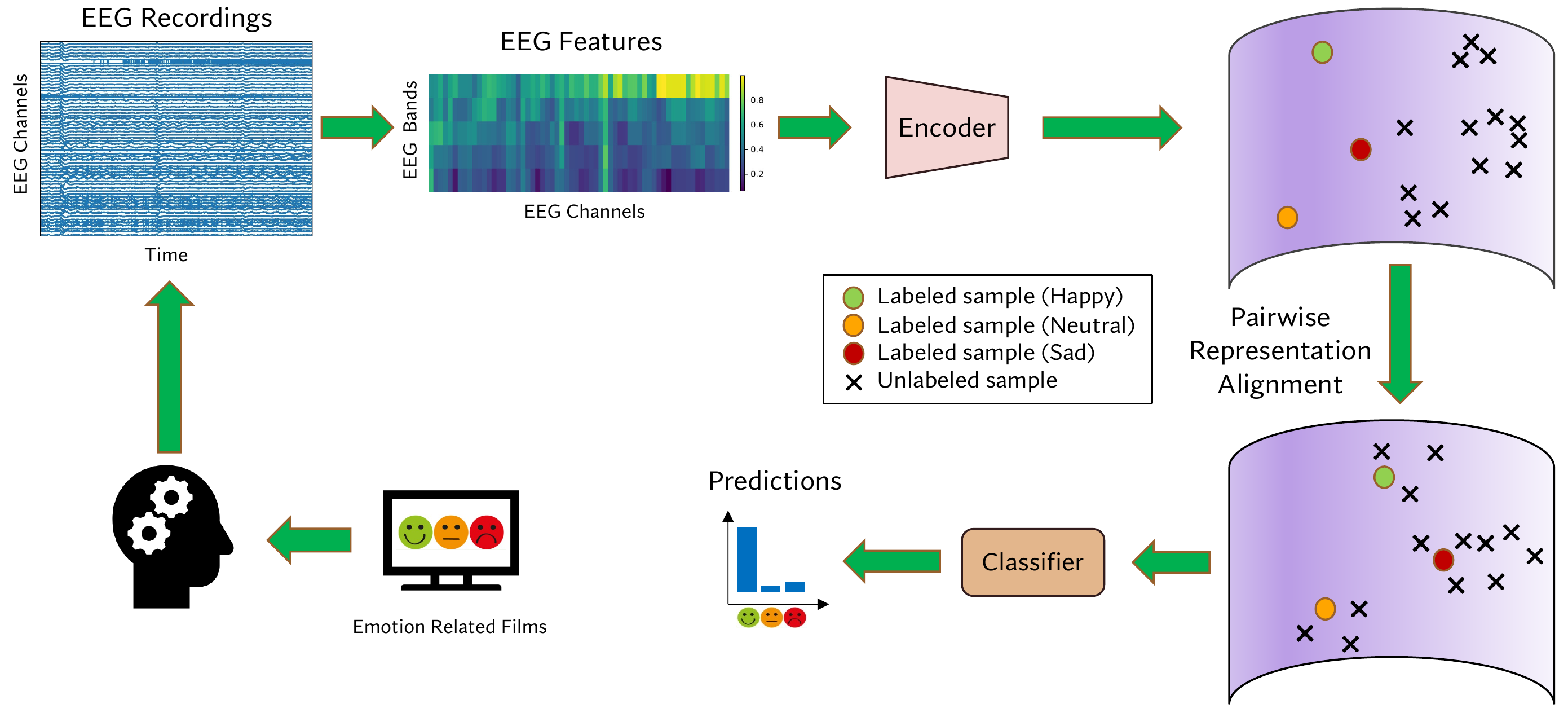} 
    \caption{An overview of our proposed semi-supervised EEG learning with pairwise representation alignment, PARSE, in the training phase is presented.}
    \label{overview}
    \end{center}
\end{figure}

\noindent \textbf{Contributions.}
In this paper, we propose a novel semi-supervised EEG learning framework for emotion recognition. Our model, entitled PARSE (\underline{P}airwise \underline{A}lignment of \underline{R}epresentations for \underline{S}emi-Supervised \underline{E}EG Learning), first augments the labeled and unlabeled EEG data. Then, we use a classifier to make predictions on the original, weakly-augmented, and strongly-augmented unlabeled data. Next, we average these three predictions as the guessed label for each unlabeled sample. Afterward, we compute convex combinations of labeled and unlabeled data followed by the representation alignment using a domain discriminator trained on the interpolated set. Simultaneously, a classifier is also trained to perform emotion recognition. An overview of our method is presented in Figure \ref{overview}. First, EEG is collected from the subject while a stimuli is provided (in the case of our study to induce different emotions). Predominant features are then extracted from the pre-processed EEG recordings. Following, an encoder is used to extract new representations. After that, to allow the classifier to predict the labels of unseen data with higher confidence, we enforce the distributions of labeled and unlabeled samples to become close to each other by aligning their representations in a pairwise manner. We test PARSE on four publicly available datasets, namely SEED \cite{zheng2015investigating}, SEED-IV \cite{zheng2018emotionmeter}, SEED-V \cite{liu2021comparing}, and AMIGOS \cite{correa2018amigos}, considering the following criteria: \textit{i}) including discrete (e.g., sad, happy, neutral) and continuous emotions (e.g., arousal); (\textit{ii}) including binary and multi-class classification tasks ($2,3,4,$ or $5$ classes); (\textit{iii}) including both balanced and imbalanced datasets. 

To fully evaluate our method, we implement and adopt several SSL methods from other domains (e.g., computer vision) to compare with our method. In particular, we implement and compare PARSE with varying amounts of few-labeled samples ($1,3,5,7,10,25$ labeled samples per class) against three cutting-edge methods, MixMatch \cite{berthelot2019mixmatch}, FixMatch \cite{sohn2020fixmatch}, and AdaMatch \cite{berthelot2021adamatch}, in addition to five classical SSL methods, $\Pi$-model \cite{samuli2017temporal}, temporal ensembling \cite{samuli2017temporal}, mean teacher \cite{tarvainen2017mean}, convolutional autoencoders \cite{tong2019caesnet}, and pseudo-labeling \cite{lee2013pseudo}. Our experiments show that PARSE achieves overall best results in SEED, SEED-IV and AMIGOS (valence), and approaches the best results in SEED-V ($0.3\%$ difference) and AMIGOS (arousal). We also perform an analysis on the impact of our pairwise representation alignment module, showing that the alignment consistently improves our method's performance over a varying number of labeled samples across all datasets. Our analysis also shows that the alignment helps to reduce the distance between the distributions of labeled and unlabeled data, particularly, when \textbf{only one} sample per class is available. This finding is further demonstrated by the effective performance of our model. 

In summary, our contributions in this paper are as follows: (\textbf{1}) We propose a novel SSL approach for EEG representation learning. Our method performs pairwise representation alignment between labeled and unlabeled samples and generalizes well across a varying number of scenarios with few labeled samples across four large public datasets.  (\textbf{2}) We perform extensive experiments and compare our method to several recent and classical SSL methods. The study shows that our approach obtains strong results, outperforming other methods in the majority of experimental conditions. We also carry out a detailed ablation study to demonstrate the impact of our pairwise representation alignment component. (\textbf{3}) To contribute to the field and enable reproducibility, we make our code publicly available at \href{https://github.com/guangyizhangbci/PARSE}{https://github.com/guangyizhangbci/PARSE}.

The rest of this paper is organized as follows. Section 2 discusses the related work in emotion recognition with EEG, semi-supervised learning, as well as semi-supervised EEG representation learning for emotion recognition. Section 3 describes our proposed semi-supervised solution for EEG learning. Section 4 presents the dataset, EEG preprocessing, and feature extraction. We also describe the evaluation protocol, implementation details, and SSL benchmarks. In section 5, we present and analyze the experiments and results, including the ablation study performed to investigate the impact of the pairwise representation alignment. Additionally, we further analyze the effect of various hyper-parameter settings on the model's performance. Finally, we summarize our work and provide a future direction and the concluding remarks.

\section{Related Work} \label{sec: related work}
\subsection{Emotion Recognition with EEG}
The general approach to emotion recognition with EEG consists of preprocessing EEG recordings, followed by feature extraction, and finally a classifier to learn the extracted features \cite{tao2020eeg,zhang2020rfnet}. In the preprocessing step, many approaches such as frequency filtering and blind signal separation are used to remove noise and artefacts from EEG data \cite{zheng2015investigating,zheng2018emotionmeter,correa2018amigos}. Following, well-known features such as logarithm Power Spectral Density (PSD) and Differential Entropy (DE) are extracted from the decomposed signals in several dominant EEG frequency bands (e.g., alpha, beta, gamma) \cite{zheng2015investigating,zheng2018emotionmeter,liu2021comparing, correa2018amigos}. Successive to feature extraction, various classification algorithms for emotion recognition have been developed \cite{zheng2015investigating,zheng2018emotionmeter,liu2021comparing, correa2018amigos}. For instance, K-nearest neighbor \cite{zheng2015investigating}, support vector machines \cite{zheng2015investigating,correa2018amigos,zheng2018emotionmeter, liu2021comparing}, logistic regression \cite{zheng2015investigating}, random forest \cite{li2020novel}, and naive Bayes \cite{correa2018amigos} have been employed to learn the non-linear output information from the extracted features. 

Recently, various deep learning techniques have been used to improve recognition performance due to their capabilities in learning more task-relevant and dominant information from extracted features or automatically learned representations. For example, deep belief networks \cite{zheng2015investigating} and deep neural networks \cite{zheng2018emotionmeter} have been used to learn higher level features using multiple hidden layers. Recurrent deep neural architectures such as long short-term memory networks \cite{zhang2016continuous} and spatial-temporal recurrent neural networks \cite{zhang2018spatial} have been used to exploit dependencies among features extracted from sequential data. A convolutional neural network framework was employed to discover spatial-temporal information \cite{zhang2021capsule} and graph neural networks were used to learn topological structure information of EEG channels through graph connections \cite{song2018eeg,zhong2020eeg}. All of the deep learning-based frameworks listed above outperform conventional machine learning classifiers, many of them achieving state-of-the-art performance in various emotion recognition tasks.

\subsection{Semi-supervised Learning}
When very few labeled data are used for training, deep learning networks may not perform well due to overfitting, slow convergence, and the random initialization of the neural networks \cite{erhan2010does}. To tackle this problem, unsupervised pre-training, for instance using Stacked Auto-Encoders (SAE) and Deep Belief Networks (DBN), has been proposed \cite{van2020survey}. In \cite{erhan2010does,van2020survey}, the unsupervised pre-training stage was claimed as a regularization and was shown to be capable of guiding the model toward a better local minima in the fine-tuning stage. This unsupervised pre-training strategy has been successfully used to improve performance on small labeled training sets in computer vision, natural language processing, and EEG learning \cite{erhan2010does,li2016unsupervised,ramachandran2017unsupervised,xu2016affective}. Recently, contrastive learning has been used for unsupervised pre-training by increasing the similarity between augmented samples of the same EEG data while decreasing that of the different ones \cite{mohsenvand2020contrastive,kostas2021bendr,banville2021uncovering}.

Pseudo-labeling \cite{lee2013pseudo} is another semi-supervised approach that encourages the model to obtain lower entropy predictions on the unlabeled data. In this method, the network (trained on labeled data) is used to generate pseudo-labels for unlabeled data, and the network is then retrained with both labeled and pseudo-labeled data.
This efficient semi-supervised paradigm outperformed many supervised techniques with small amounts of labeled training samples \cite{lee2013pseudo}.

Consistency regularization claims that a label should remain consistent even after a perturbation or augmentation is applied on its data, which has been widely used in semi-supervised learning \cite{samuli2017temporal,tarvainen2017mean,oliver2018realistic}. The most common regularization approaches that can be applied on both unlabeled and labeled data are stochastic augmentations (e.g., Gaussian noise) imposed on inputs and a dropout layer applied in the network. $\Pi$-model \cite{samuli2017temporal} encourages the network to give consistent outputs for two augmentations on the same input. Temporal ensembling \cite{samuli2017temporal} uses a similar consistency regularization technique and aggregates the network predictions from previous training epochs. Mean teacher \cite{tarvainen2017mean} is also built based on the $\Pi$-model and works by aggregating the network weights from previous training batches.

Recently, a few complex pipelines, also known as `holistic' approaches, have been proposed by combining various elements of the SSL ideas described above. For example, MixMatch \cite{berthelot2019mixmatch} utilizes consistency regularization and label-entropy minimization based on several techniques such as MixUp \cite{zhang2018mixup} and prediction sharpening. Compared to MixMatch, FixMatch \cite{sohn2020fixmatch} provides a simplified solution by applying a customized threshold for choosing high quality pseudo-labels. AdaMatch \cite{berthelot2021adamatch} encourages the label distribution of predictions on unlabeled data to become closer to that of labeled data. It also uses a dynamic threshold based on the network's confidence on predictions of labeled data to choose high quality pseudo-labels for unlabeled data.

\subsection{Semi-supervised EEG Emotion Recognition}
Semi-supervised learning has been very rarely studied in the context of EEG-based emotion recognition. Recently, in \cite{zhang2021deep}, we proposed a deep semi-supervised architecture with an attention-based recurrent autoencoder for EEG learning. We also re-implemented several popular SSL pipelines equipped with deep learning techniques, namely unsupervised pre-training \cite{erhan2010does}, pseudo-labeling \cite{lee2013pseudo}, $\Pi$-model \cite{samuli2017temporal}, temporal ensembling \cite{samuli2017temporal}, and mean teacher \cite{tarvainen2017mean}, which were all originally proposed in the field of computer vision. We compared our framework to the aforementioned SSL methods on a large-scale EEG emotion dataset, SEED. Very lately, in \cite{zhang2021holistic}, we adapted and implemented three recent state-of-the-art holistic SSL methods, MixMatch \cite{berthelot2019mixmatch}, FixMatch \cite{sohn2020fixmatch}, and AdaMatch \cite{berthelot2021adamatch} for EEG representation learning. We also compared these three holistic SSL techniques to the other five classical SSL methods, $\Pi$-model \cite{samuli2017temporal}, temporal ensembling \cite{samuli2017temporal}, mean teacher \cite{tarvainen2017mean}, convolutional autoencoder \cite{tong2019caesnet}, and pseudo-labeling \cite{lee2013pseudo} on two popular emotion recognition datasets, SEED and SEED-IV. Our study demonstrated the great potential for semi-supervised EEG-based emotion recognition, and motivated this study to develop novel semi-supervised methods designed for the field of EEG representation learning.

\section{Proposed Solution} \label{sec: proposed solution}
\subsection{Problem Setup}
For a classification problem with $k$ emotion categories, let us denote $\mathcal{D}_l=\{(x_i^l, y_i^l)\}_{i=1}^M$, $\mathcal{D}_u= \{x_i^u\}_{i=1}^N$, $\mathcal{D}_v=\{(x_i^l, y_i^l)\}_{i=1}^I$, and $\mathcal{D}$ as the labeled, unlabeled, validation, and overall training sets, where $\mathcal{D}_l \cup \mathcal{D}_u \cup \mathcal{D}_v  = \mathcal{D}$ and $\mathcal{D}_l \cap \mathcal{D}_u \cap \mathcal{D}_v= \emptyset$. $\mathcal{D}_l$ is formed by $m$ samples per class selected from $\mathcal{D}$ ($M=m\times k$). Our goal is to propose a robust pipeline to improve the model's performance on emotion recognition by leveraging large amounts of $\mathcal{D}_u$ when limited $\mathcal{D}_l$ are available. We are interested in multiple few-labeled scenarios when $M\ll N$, including a barely supervised scenario, where \textbf{only one} sample per class is labeled ($m=1$). 

\subsection{Solution Overview}
\begin{figure*}
    \begin{center}
    \includegraphics[width=2.0\columnwidth]{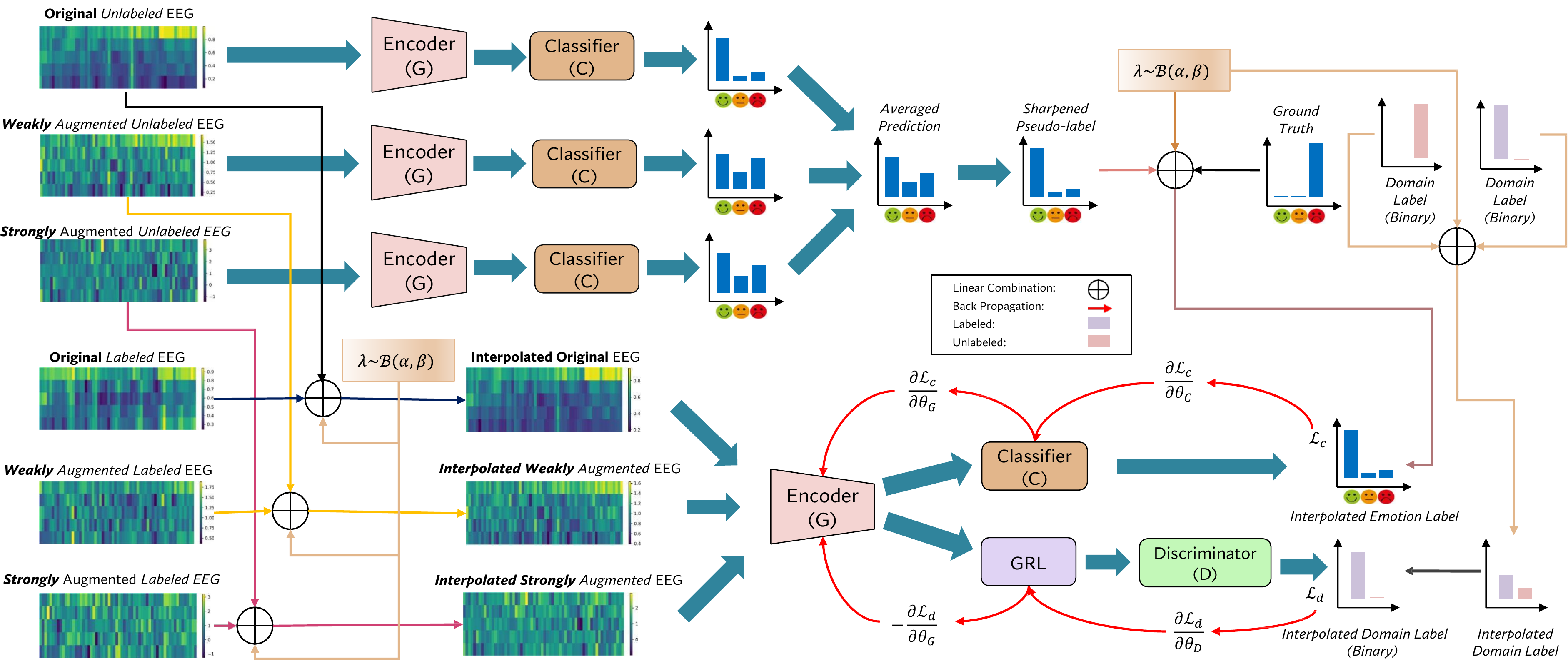} 
    \caption{The architecture of our proposed semi-supervised framework with pairwise representation alignment, PARSE, for emotion recognition.}
    \label{architecture}
    \end{center}
\end{figure*}

We aim to propose a novel SSL architecture for EEG-based emotion recognition, while reducing the distribution mismatch between large unlabeled and small labeled data. To achieve this, we first apply two types of augmentations (strong and weak) on both labeled and unlabeled EEG data. Following that, we compute the model's averaged predictions on both the original and augmented unlabeled data. To reduce the entropy of the averaged predictions, we further employ a sharpening process and use the sharpened prediction as guessed labels for all the unlabeled data. Next, we use convex combinations of pairwise labeled and unlabeled data to form a new set. After that, we simultaneously perform emotion recognition and pairwise representation alignment by training an emotion classifier and a domain discriminator (\textit{labeled} vs. \textit{unlabeled}) on the new interpolated set. The overview of our proposed solution is shown in Figure \ref{architecture}. In the following subsections, we describe each step for our proposed method in details.

\subsection{Our Approach}\label{sec: our approach}
In order to efficiently train our model using the entire large set of unlabeled data ($\mathcal{D}_u$) together with the small amounts of labeled data ($\mathcal{D}_l$), we first replicate $\mathcal{D}_l$ by $\floor{N/M}$ times to make it the same size as $\mathcal{D}_u$. In each training iteration, we randomly select batches of samples $(x_b^l, y_b^l) \in \mathcal{D}_l$ from the labeled dataset and $(x_b^u) \in \mathcal{D}_u$ from the unlabeled set, respectively. Next, we perform data augmentation on each data batch as described below.

\subsubsection{Data Augmentation}
Data augmentation is an important step in semi-supervised learning. In fundamental SSL literature \cite{berthelot2019mixmatch,sohn2020fixmatch,berthelot2021adamatch,ghosh2020data}, data augmentation methods generally include (1) geometric transformations, and (2) noise addition. Geometric transformations, such as flipping, rotation, and scaling, are not suitable for time-series data such as EEG \cite{wang2018data}. Possible types of added noise include salt-and-pepper, Poisson, and Gaussian \cite{wang2018data}. Salt-and-pepper and Poisson noise are not recommended for EEG data augmentation since they might alter the EEG's intrinsic features \cite{wang2018data,lashgari2020data}. Therefore, we adopt additive Gaussian noise which has been widely used for data augmentation in both recent EEG studies \cite{mohsenvand2020contrastive,wang2018data,li2019channel,lashgari2020data,luo2020data} and other domains \cite{berthelot2019mixmatch,ghosh2020data}.

We first apply strong and weak augmentations on both labeled and unlabeled data. For each $x_b^l$ and $x_b^u$ in the training set of $D_l$ and $D_u$, we generate the augmented data as $\mathcal{A}_{s/w}(x_b) = x_b + \mathcal{N}(\mu, \sigma)$, where $x_b\sim[0,1]$ is the normalized input, and $\mathcal{N}$ is a Gaussian distribution with $\mu=0.5$ which is consistent with the mean value of the normalized input. The strength of the augmentation can be tuned by changing $\sigma$. We choose $0.8$ and $0.2$ as $\sigma$ in the additive Gaussian noise for strong ($\mathcal{A}_{s}$) and weak ($\mathcal{A}_{w}$) augmentations respectively, as suggested in \cite{liu2019multimodal,zhang2021capsule}. Following data augmentation, label-guessing is required to obtain pseudo-labels for both the original and augmented unlabeled data, which we describe below.

\subsubsection{Label-Guessing} \label{label-guessing}
After data augmentation, we enlarge the unlabeled set of data by concatenating the original ($x_b^u$), weakly ($\mathcal{A}_w(x_b^u)$), and strongly ($\mathcal{A}_s(x_b^u)$) augmented unlabeled samples. We then pass-forward the unlabeled data to an encoder ($G$) and a classifier ($C$) to generate pseudo-labels ($p_b$) by averaging the predictions as: 
\begin{equation}\label{average_prediction}
p_b = \frac{1}{3}\sum_1^{3} \softmax(p_m(y\mid x_b^{u,r}; \theta_G, \theta_C)),
\end{equation} 
where $x_b^{u,r}$ for $r=1,2,3$, denotes $x_b^u$, $A_w(x_b^u)$, and $A_s(x_b^u)$, respectively.
Here, $p_m$ represents the model prediction, and $\theta_G$ and $\theta_C$ denote the model's parameters for the encoder and classifier, respectively. Then, to minimize the entropy of the pseudo-labels, we apply a sharpening operation by:
\begin{equation}\label{sharpen}
p_b^u = p_b^{1/T} \Big/\sum_1^k(p_b^{1/T}),
\end{equation} 
where $T=1$ is the temperature hyper-parameter used to adjust the entropy level of the pseudo-labels ($p_b$), as suggested in \cite{berthelot2019mixmatch}. 

In addition to emotion pseudo-labels, we also assign binary domain (labeled Vs. unlabeled) labels ($z^u$) to the unlabeled set ($G_u$) as:  
\begin{equation}\label{new_unlabeled_set}
G_u = \{(\langle x_b^u , \mathcal{A}_s(x_b^u) , \mathcal{A}_w(x_b^u)\rangle, \langle p_b^u , p_b^u , p_b^u \rangle, z^u)\},
\end{equation} 
where $\langle , \rangle$ denotes the concatenation of two or more sets. 

Similarly, we enlarge the labeled data with the concatenation of the original ($x_b^l$), weakly ($\mathcal{A}_w(x_b^l)$), and strongly ($\mathcal{A}_s(x_b^l)$) augmented labeled samples. Since the augmentation should not alter the labels of the data, we employ the ground truth of the original labeled data as labels for the weakly- and strongly-augmented labeled data. Moreover, similar to the unlabeled set ($G_u$), we also assign binary domain labels ($z^l$) to the labeled set ($G_l$) as:
\begin{equation}\label{new_labeled_set}
G_l = \{(\langle x_b^l , \mathcal{A}_s(x_b^l) , \mathcal{A}_w(x_b^l)\rangle, \langle y_b^l , y_b^l , y_b^l\rangle, z^l)\},
\end{equation} 
where $z^l \neq z^u$ and $|z^l|=|z^u|=3|y_b^l|$ ($|.|$ denotes the size of a set). The data and corresponding labels from both $G_l$ and $G_u$ are then used for calculating pairwise convex combinations, which will be discussed in the following.

\subsubsection{MixUp}\label{sec:mixup}
The generalization performance of a model could degrade due to the adversarial examples whose model predictions could be easily altered by small perturbations (e.g., Gaussian noise used for our data augmentations) \cite{verma2019interpolation,zhang2020does}. To tackle this problem, we encourage using convex combinations between pairs of samples, as inspired by a powerful method called MixUp \cite{zhang2018mixup}. Specifically, we generate a new training set by using the convex combinations of the \textit{labeled} and \textit{unlabeled} set ($G_l, G_u$), according to: 
\begin{equation}\label{beta_distribution}
\lambda \sim \mathcal{B}(\alpha, \beta),
\end{equation}
\begin{equation}\label{mixup_x}
 \tilde{x}_b = \lambda \langle x_b^l , \mathcal{A}_s(x_b^l) , \mathcal{A}_w(x_b^l)\rangle + (1 - \lambda)\langle x_b^u , \mathcal{A}_s(x_b^u) , \mathcal{A}_w(x_b^u)\rangle,
\end{equation}
\begin{equation}\label{mixup_y}
 \tilde{y}_b = \lambda \langle y_b^l , y_b^l , y_b^l\rangle + (1 - \lambda) \langle p_b^u , p_b^u , p_b^u \rangle,
\end{equation} 
\begin{equation}\label{mixup_z}
 \tilde{z}_b = \lambda z_b^l + (1 - \lambda) z_b^u,
\end{equation} where $\mathcal{B}$ is the Beta distribution with $\alpha, \beta$ of $0.25$, as suggested in \cite{zhang2018mixup,berthelot2019mixmatch,wang2019semi}, and $\lambda \in [0,1]$. This matching is performed between randomly selected pairs of the labeled and unlabeled samples. We determine the binary domain label through a rounding operation applied on Eq. \ref{mixup_z} as $\nint{\tilde{z}_b}.$
Finally, a new combined set containing triplets of ($\tilde{x}_b, \tilde{y}_b, \nint{\tilde{z}_b}$) will be used in our pairwise representation alignment step.

\subsubsection{Pairwise Representation Alignment}
Following MixUp, we align the distribution of the pairwise labeled and unlabeled embeddings, that are obtained by training the encoder using the newly interpolated triplet of ($\tilde{x}_b, \tilde{y}_b, \nint{\tilde{z}_b}$). To measure the distribution divergence between the labeled and unlabeled data, we use:
\begin{equation}\label{divergence}
d_H = 2\{1-\min_H[\frac{1}{|\tilde{x}_b|}\sum_i^{|\tilde{x}_b|}\argmax (p_m(z\mid x_i; \theta_G, \theta_D))\neq z_i]\}, 
\end{equation} 
based on \cite{ganin2016domain,wang2019semi}, in which $x_i \in \tilde{x}_b$ and $z_i \in \nint{\tilde{z}_b}$. $\theta_D$ is the model parameters for the discriminator ($D$). The min() operation is applied on the discriminator's prediction error (labeled vs. unlabeled) so that distribution divergence would be larger when the prediction error is smaller. Parameters $\theta_G$ and $\theta_D$ need to be optimized during this operation. We minimize this distribution divergence in order to encourage the encoder ($G$) to align the EEG representations of labeled and unlabeled data as: 
\begin{equation}\label{max_min}
\small
\min_G d_H = \max_G \min_H[\frac{1}{|\tilde{x}_b|}\sum_i^{|\tilde{x}_b|}\argmax (p_m(z\mid x_i; \theta_G, \theta_D))\neq z_i]\}.
\end{equation}

As suggested in \cite{ganin2016domain,wang2019semi}, we optimize the max-min problem by adding a Gradient Reverse Layer (GRL) before the $D$ in order to reverse the gradient in $G$, as shown in Figure \ref{architecture}. Specifically, GRL is used to reverse the sign of the gradients during back-propagation so that the discriminator loss can be directly minimized using the existing optimization algorithms (e.g., Adam \cite{kingma2014adam}) \cite{zhang2018collaborative}. 
Finally, we train both the emotion classifier and the domain discriminator by minimizing the emotion classification loss: 
\begin{equation}\label{loss_c}
\mathcal{L}_c = ||\tilde{y}_b- \softmax(p_m(y \mid \tilde{x}_b;\theta_G,\theta_C))||_2^2,
\end{equation} 
and domain discriminator loss:
\begin{equation}\label{loss_d}
\mathcal{L}_d = \mathcal{H}(\nint{\tilde{z}_b}, p_m(z\mid \tilde{x}_b; \theta_G, \theta_D)),
\end{equation}
where $\mathcal{H}(p, q) = -\sum p(x)\log q(x)$ represents cross-entropy of $p$ and $q$. We use an adversarial training strategy to minimize the distance between the labeled and unlabeled representations.

\subsubsection{Total Loss Function}
Our total loss function comprises three parts. The first term is a supervised loss $\mathcal{L}_s = \mathcal{H}(y_b^l, p_m(y\mid x_b^l; \theta_G, \theta_C))$ that has been commonly used in many SSL literature. We adopt the unsupervised loss $L_u$ used in \cite{sohn2020fixmatch}, as the second term as follows:
\begin{equation}\label{l_u}
\mathcal{L}_u = \mathcal{H}(y_b^u, p_m(y\mid \mathcal{A}_s(x_b^u);\theta_G, \theta_C)),
\end{equation} where $y_b^u = \argmax(p_m(y\mid \mathcal{A}_w(x_b^u);\theta_G, \theta_C))$. Consequently, we update the total loss as:
\begin{equation}\label{total_loss}
\mathcal{L}_{total} = \mathcal{L}_s+ \eta\mathcal{L}_u + \delta(\lambda \mathcal{L}_{c}+\mathcal{L}_{d}),
\end{equation}
where $\delta=1.0$. Instead of using a pre-defined threshold $\tau$ \cite{sohn2020fixmatch} that may need to be tuned for each dataset individually, we apply a warm-up function $\eta$ on the unsupervised loss, similar to \cite{berthelot2021adamatch}, as follows:
\begin{equation}\label{warm_up}
\eta(t) = \frac{1}{2} - \cos(\min(\pi, 2\pi t/T))/2, 
\end{equation} 
where $t$ and $T$ are the current and the maximum iterations. The warm-up function is used to slowly increase the weight of the unsupervised loss when the model is being trained further as well as being more confident on its predictions. The third term is the sum of the classification and discriminator losses trained on the interpolated set as mentioned earlier.

\subsection{Architecture Details} \label{sec: arch details}
Here we describe the architecture details of the different modules, namely encoder, classifier, and discriminator, used in our proposed method, as shown in Table \ref{model}. The encoder consists of two $1$-D convolutional blocks, where each block contains a $1$-D convolutional layer followed by a $1$-D batch normalization layer and a LeakyReLU activation. The classifier and discriminator share the same architecture containing two fully connected layers with a dropout rate of $0.5$. The encoder is used to transform EEG inputs into a learned embedding, while the classifier is used for identifying emotion categories, and the discriminator is used to determine whether the input data are labeled or unlabeled. In Table \ref{model}, $s$ denotes the total number of EEG features, and $k$ is the number of emotion categories.

\section{Experimental Setup and Settings}
\subsection{Datasets}
We use the following four datasets in our study. In all the dataset mentioned below, the EEG electrodes were placed using $10-20$ system. 

\subsubsection{SEED}
The SEED dataset was built by Zheng and Lu \cite{zheng2015investigating}. $15$ film clips with three emotions (neutral, positive and negative) were chosen and used as stimuli in the experiments. The studies were conducted by a total of $15$ participants, consisting of $8$ females and $7$ males. Each participant performed the experiment twice, with each experiment consisting of $15$ trials. Each trial has a $4$-step pipeline: $5$ seconds of start hint before the film clip, $4$ minutes of the clip as emotion stimulus, $45$ seconds of self-assessment, and finally $15$ seconds of break. EEG was recorded from $62$ electrodes at a sampling rate of $1000$\textit{Hz}.

\begin{table}[t]
\caption{Architectural details of our proposed model.}
\tiny
\label{model}
\resizebox{\columnwidth}{!}
{
\begin{tabular}{c|c|c}
\hline
\textbf{Module} & \textbf{Layer details}  & \textbf{Output shape} \\[.1cm] \hline
\textbf{Input} & -  & $(1, s)$ \\[.1cm] \hline
\multirow{4}{*}{\textbf{Encoder}} & \convMatrix{3}{5}{5} & $(5, s-2)$ \\[.1cm] 

\cline{2-3} & \convMatrix{3}{10}{10} & $(10, s-4)$ \\[.1cm]
\cline{2-3} \hline \textbf{Embedding} & Flatten & $10\times (s-4)$ \\[.1cm]
\cline{2-3}
\hline
\multirow{2}{*}{\textbf{Classifier}} & \convMatrixc{k} & $(k)$ \\[.1cm] 
    
\hline
\multirow{2}{*}{\textbf{Discriminator}} & \convMatrixc{2} & $(2)$ \\[.1cm] 
\hline

\end{tabular}
}
\end{table}

\subsubsection{SEED-IV}
Zheng et al. developed the SEED-IV dataset, which was initially used in \cite{zheng2018emotionmeter}. As stimuli, $72$ film clips with four emotions (sad, fear, happy and neutral) were selected. The experiments were completed by $15$ individuals, consisting of $8$ females and $7$ males. Each participant repeated the experiment three times, with completely different stimuli each time. Each experiment has a total of $24$ trials ($6$ trials for each emotion), where each trial has three stages: $5$ seconds of start hint, $2$ minutes of film clip, followed by $45$ seconds of self-assessment. $62$ EEG recordings were collected at a sampling frequency of $1000$\textit{Hz}.

\subsubsection{SEED-V}
The SEED-V dataset was collected by Liu et al. \cite{liu2021comparing}. $45$ short videos with five emotions (happy, fear, neutral, sad, and disgust) were chosen as stimuli. The studies were completed by a total of $16$ participants, $10$ females and $6$ males. Each participant repeated the experiment three times, with completely different stimuli each time. Each experiment includes $15$ trials ($3$ trials for each emotion), where each trial has three stages: $15$ seconds of start hint, $2-4$ minutes of film clip, and $15$ or $30$ seconds of self-assessment. In total, $62$ EEG recordings were collected with a sampling frequency of $1000$\textit{Hz}.

\subsubsection{AMIGOS}
The AMIGOS dataset was developed by Correa et al. \cite{correa2018amigos}. $37$ participants ($12$ females and $25$ males) completed experiments with both $16$ short and $4$ long video clips as stimuli. Experiments with short video stimuli consist of three stages: $5$ seconds of baseline recording, a few minutes of video clips of varying duration (less than $250$ seconds), and self-assessment of arousal, valence as well as other affective states. Experiments with long video stimuli also consist of three stages: $45$ seconds of initial self-assessment of several affective states including arousal and valence, two video clips of varying lengths (more than $14$ minutes), and another $45$ seconds of self-assessment. Videos of participants during experiments were cropped to display the face and then segmented into continuous $20$-second clips. Three annotators assigned valence and arousal scores (between $[-1, 1]$) to each of these $20$-second video snippets. A threshold of ($0.0$) was used to transform continuous scores into binary classes (positive and negative). $14$ EEG channels were recorded at a sampling rate of $128$\textit{Hz}.

\subsection{Preprocessing}
EEG preprocessing is often dataset-dependent due to the various collection equipment, as well as experimental conditions and protocols. It often contains downsampling, artefact removal, and noise filtering. In our experiments, we use the \textit{preprocessed} data that are made public by the original papers which introduced the datasets \cite{zheng2015investigating,zheng2018emotionmeter,liu2021comparing,correa2018amigos}, and do not apply any further preprocessing.

\subsubsection{Downsampling} 
In all three SEED-series datasets, EEG signals were downsampled from $1000$ \textit{Hz} to $200$ \textit{Hz}. In AMIGOS dataset, EEG data were already recorded at a low sampling rate of $128$ \textit{Hz} and were not further downsampled to avoid aliasing effect. 

\subsubsection{Artefact Removal and Noise Filtering} 
EEG signals are often contaminated by Electrooculogram (EOG) and Electromyogram (EMG) \cite{zheng2015investigating,katsigiannis2017dreamer}. EOG signals, which reflect the ocular activities such as eye movements and eye blinks, are most active below $4$\textit{Hz} \cite{katsigiannis2017dreamer}. EMG signals reflect the muscular activities around the face and are dominant above $30$\textit{Hz} \cite{katsigiannis2017dreamer}. Both EOG and EMG have a high overlap with EEG, which is dominant in $0.3-50$\textit{Hz} \cite{zheng2015investigating}. 
Consequently, different artefact removal and noise filtering methods were used in the original papers that published the datasets. To minimize the artefacts, in SEED, the EEG signals were visually checked by experts, and the recordings identified as highly contaminated by EOG and EMG were discarded \cite{zheng2015investigating}. 
In SEED-V, principle component analysis was used to remove ocular artefacts from EEG by \cite{zhao2019classification,liu2021comparing}. Similarly in AMIGOS, blind source separation was employed to remove the eye artefact \cite{correa2018amigos}. At last, a band-pass filter was applied to filter the noise and artefacts outside the dominant EEG frequency range. Specifically, the frequency ranges of the bandpass filter are $0.3-50$\textit{Hz}, $1-75$\textit{Hz}, $1-75$\textit{Hz}, and $4-45$\textit{Hz} for SEED, SEED-IV, SEED-V, and AMIGOS datasets, respectively \cite{zheng2015investigating,zheng2018emotionmeter,liu2021comparing,correa2018amigos}. We use the preprocessed signals made public by the original publications \cite{zheng2015investigating,zheng2018emotionmeter,liu2021comparing,correa2018amigos} and do not perform further preprocessing.

\subsection{Feature Extraction}
Following preprocessing, we use pre-defined EEG segments proposed in the respective dataset publications \cite{zheng2015investigating,zheng2018emotionmeter,liu2021comparing,correa2018amigos}, for feature extraction. In each of the SEED-series datasets, EEG data were divided into segments of the same length with no overlap between adjacent segments. The lengths of EEG segments were set to $1$ second for SEED and $4$ seconds for SEED-IV and SEED-V, respectively \cite{zheng2015investigating,zheng2018emotionmeter,liu2021comparing}. For AMIGOS, EEG signals corresponding to each video clip were separated into $20$-second segments. The first and last segments were selected from the EEG data corresponding to the initial and final $20$ seconds of the video clips. Next, following the first $5$ seconds of video clips, EEG data were divided into non-overlapping $20$-second segments, yielding $340$ EEG segments for each participant \cite{correa2018amigos}. 
It should be noted that the extracted features (described below) are reshaped from 2-D to 1-D feature vectors (e.g., [$62$ channels, $5$ features/channel] reshaped to [$310\times 1$ features]) and further scaled into the range of $[0, 1]$ with min-max normalization before being fed to our proposed framework. The 1-D feature vector begins with features extracted from each frequency band (e.g., delta, theta, alpha, beta, and gamma) of the first EEG channel, and continues with features extracted in the same order from the other EEG channels.

\subsubsection{Differential Entropy} \label{sec: differential_entropy}
In SEED-series datasets, DE features were extracted from five EEG bands, notably delta ($1-4$\textit{Hz}), theta ($4-8$\textit{Hz}), alpha ($8-14$\textit{Hz}), beta ($14-31$\textit{Hz}), and gamma ($31-50$\textit{Hz}). Consequently, a total of $62\times5=310$ features were extracted. We assume that the signals have a Gaussian distribution, and thus DE is calculated as follows:
\begin{equation}\label{equation: de}
DE = \frac{1}{2} \log{2\pi e \sigma^{2}}.  
\end{equation}

\subsubsection{Power Spectral Density}
In AMIGOS dataset, the logarithm of PSD features were computed from five EEG bands, namely theta ($3-7$\textit{Hz}), slow alpha ($8-10$\textit{Hz}), alpha ($8-13$\textit{Hz}), beta ($14-29$\textit{Hz}), and gamma ($30-47$\textit{Hz}). In addition, the logarithm of PSD asymmetry between $7$ symmetric pairs of EEG channels (e.g., F$7$ and F$8$) were also extracted as features. As a result, a total of $(14+7)\times5=105$ features were extracted. The PSD is calculated as:
\begin{equation}
S_{xx}(\omega)=\lim_{T\to\infty}E\Big[|\hat{X}(\omega)|^{2}\Big].
\end{equation}

\subsection{Evaluation Protocols}
To evaluate our proposed pipeline, we strictly follow the same evaluation protocols used for emotion recognition in the original articles that published the datasets \cite{zheng2015investigating,zheng2018emotionmeter,liu2021comparing,correa2018amigos}. In SEED, we use the first $9$ trials as training data (each emotion class with three trials) and the rest $6$ trials as the testing data in each experiment, as defined in \cite{zheng2015investigating}. In SEED-IV, we employ the pre-defined first $16$ trials (each emotion class with four trials) as the training set, and the remaining $8$ trials as the testing set. In SEED-V, $15$ trials were split into three pre-defined groups, each containing $5$ trials with all $5$ emotions. We concatenate the first group from each of the three experiments to form a new fold of $15$ trials. Similarly, we form the other two folds by concatenating the second and third groups from each of the three experiments. Following that, we perform a $3$-fold cross-validation as performed in \cite{liu2021comparing}. Since the class distributions are almost balanced in the SEED-series datasets, accuracy is selected as the evaluation metric \cite{zheng2015investigating,zheng2018emotionmeter,liu2021comparing}. In AMIGOS, we adopt the same leave-one-participant-out protocol for training and testing data splits, as used in \cite{correa2018amigos}. The ratios of negative/positive classes are $0.721/0.279$ and $0.805/0.195$ for valence and arousal, respectively \cite{correa2018amigos,shukla2019feature}. As the class distribution is very imbalanced, we use the F$1$-score (mean F$1$-score for both classes) as the evaluation metric as suggested in \cite{correa2018amigos}.

\subsection{Implementation Details}\label{sec: implementation}
We use a batch size of $8$ for the datasets with subject-dependent evaluation (SEED, SEED-IV and SEED-V) and $64$ for the dataset with subject-independent evaluation (AMIGOS). We adopt the Adam algorithm \cite{kingma2014adam} with a default learning rate of $1e^{-3}$ for optimization.
We train our model for a total of $30$ epochs for all the experiments. We implemented our experiments using PyTorch on a pair of NVIDIA GeForce RTX $2080$ Ti GPUs.

Similar to other semi-supervised studies \cite{berthelot2019mixmatch,sohn2020fixmatch,berthelot2021adamatch}, we evaluate PARSE with a varying number of labeled samples per class ($m$), where $m \in \{1,3,5,7,10, 25\}$. For the benchmark methods using a dataset-specific hyper-parameter $\tau$, we perform a smart search in the range of $[0.0-1.0]$ with a step size of $0.1$ on the validation set (not the test set) to find the optimum hyper-parameter ($\tau$). Specifically, in FixMatch, we set $\tau = 0.9$ for SEED and SEED-IV, $\tau = 0.7$ for SEED-V and $\tau = 0.6$ for AMIGOS. In AdaMatch, we set $\tau = 0.6$ for SEED and AMIGOS, $\tau = 0.5$ for SEED-IV, as well as $\tau = 0.9$ for SEED-V.

\subsection{SSL Benchmarks} \label{section: ssl benchmarks}
For comparison and evaluation of our work, we adapt, implement, and necessary modify three cutting-edge SSL methods, MixMatch \cite{berthelot2019mixmatch}, FixMatch \cite{sohn2020fixmatch}, and AdaMatch \cite{berthelot2021adamatch}, in addition to five classical SSL methods, $\Pi$-model \cite{samuli2017temporal}, temporal ensembling \cite{samuli2017temporal}, mean teacher \cite{tarvainen2017mean}, convolutional autoencoders \cite{tong2019caesnet}, and pseudo-labeling \cite{lee2013pseudo}, for EEG representation learning. We implement the aforementioned $\Pi$-model, temporal ensembling, mean teacher, pseudo-labeling, and convolutional autoencoder with the same algorithm settings (e.g., loss function, unsupervised loss coefficient, etc.) used in \cite{zhang2021deep}. Lastly for MixMatch, FixMatch, and AdaMatch, we follow the implementation and hyper-parameter details presented in \cite{zhang2021holistic}.

\subsubsection{$\Pi$-model}
$\Pi$-model, as proposed in \cite{samuli2017temporal}, first applies two different augmentations on the same input. It then trains the network with dropout on the two augmented inputs, and enforces an unsupervised consistency by reducing the distance between the two corresponding outputs of the network. The supervised loss is simply the cross-entropy calculated between labeled data and the ground truth. $\Pi$-model outperformed pseudo-labeling in multiple datasets with small amounts of labeled data \cite{oliver2018realistic}. $\Pi$-model, however, trains slowly because it trains on each input twice. Furthermore, because of the single evaluation in each training epoch, the network outputs may be noisy or unreliable \cite{samuli2017temporal}.

\subsubsection{Temporal Ensembling} \label{sec: temporal ensembling}
Temporal ensembling was proposed as an improved version of the $\Pi$-model by training the network only once on each input, and aggregating the predictions from earlier training epochs \cite{samuli2017temporal}. As a result, compared to the $\Pi$-model, temporal ensembling requires less training time. Moreover, consistency regularization is applied between the current network output and the aggregated output, which is more reliable \cite{samuli2017temporal,oliver2018realistic}. However, due to the slow update from aggregating the network predictions (once per training epoch), temporal ensembling may not be able to provide promising results when the dataset is too large \cite{tarvainen2017mean}.

\subsubsection{Mean Teacher}  \label{sec: mean teacher}
In \cite{tarvainen2017mean}, a more enhanced version of temporal ensembling, called mean teacher, was proposed. Mean teacher method updates more frequently by aggregating the model weights from each previous training batch instead of ensembling the model predictions from past training epochs. Afterward, the consistency regularization is applied by minimizing the distance between the two outputs of the network with and without the ensembled model weights. Mean teacher method has obtained better results than the $\Pi$-model and temporal ensembling in a few datasets \cite{tarvainen2017mean,oliver2018realistic}. 

\subsubsection{Pseudo-Labeling}   \label{sec: pseudo labeling}
In order to generate pseudo-labels, first the model is trained using labeled data. Then, the trained model is used to obtain confident predictions on the unlabeled samples (also called pseudo-labels). Finally, the model is retrained using the entire data with true labels and pseudo-labels together \cite{lee2013pseudo}.

\subsubsection{Convolutional autoencoder}
We employ the same convolutional encoder and classifier in these benchmarks as in the one used in our proposed method. For the decoder component of the convolutional autoencoder benchmark, we use two transposed convolutional $1$-D blocks. In each block, a $1$-D transposed convolutional layer is followed by a $1$-D batch normalization layer and ReLU activation. In the pre-training stage, the convolutional autoencoder is trained on the unlabeled data to update the weights of each layer and capture a better latent representation of the inputs while minimizing the unsupervised loss. Then, the pre-trained network (e.g., encoder) followed by a classifier is fine-tuned using labeled data in a supervised manner \cite{van2020survey}.

\subsubsection{MixMatch}   \label{sec: mixmatch}
MixMatch \cite{berthelot2019mixmatch} has been built upon two SSL core concepts, namely consistency regularization applied in \cite{samuli2017temporal,tarvainen2017mean,oliver2018realistic} and label entropy minimization used in pseudo-labeling \cite{lee2013pseudo}. Specifically, in the first step, data augmentations are applied to both labeled and unlabeled data. In the second step, the augmented unlabeled data are fed forward to a model to obtain the predictions, with no gradients propagated during this step. Following this, the predictions are averaged to obtain the guessed labels. The entropy of the guessed labels' distributions is then minimized using a sharpening method. In the third step, a powerful data augmentation method, named MixUp \cite{zhang2018mixup}, is employed to combine the augmented labeled data with ground truth and the augmented unlabeled data with the guessed labels into a new set \cite{berthelot2019mixmatch}. In the last step, the model is trained by minimizing the cross-entropy loss between the predictions on the labeled data and ground truth, as well as minimizing the $L2$ loss between the predictions on the unlabeled data and the guessed labels of the new mixed set. MixMatch outperformed $\Pi$-model, mean-teacher, and pseudo-labeling in multiple datasets given a few labeled examples \cite{berthelot2019mixmatch}.

\subsubsection{FixMatch}   \label{sec: fixmatch}
FixMatch, a more accurate SSL technique, was proposed in \cite{sohn2020fixmatch}. In comparison to MixMatch \cite{berthelot2019mixmatch}, FixMatch combines consistency regularization and pseudo-labeling in a much simpler manner. Specifically, FixMatch first applies weak and strong augmentations to unlabeled data, and a weak augmentation to labeled data. Then, the weakly-augmented unlabeled data are fed to a model to obtain pseudo-labels. Next, the model is trained on strongly-augmented unlabeled data by minimizing cross-entropy loss between the model's prediction and the pseudo-labels, rather than reducing their squared difference which was used in the previous SSL methods \cite{samuli2017temporal,tarvainen2017mean,oliver2018realistic,berthelot2019mixmatch} as the unsupervised loss. During training, a user-defined threshold is employed to ensure that pseudo-labels are only used for unsupervised loss update when the model is confident in its predictions on the unlabeled data. Additionally, the cross-entropy loss between model's predictions on weakly-augmented labeled data and the ground truth is used for the supervised term. FixMatch achieved better performance than the related works discussed above when only a small number of samples were labeled \cite{sohn2020fixmatch}.

\subsubsection{AdaMatch}   \label{sec: adamatch}
AdaMatch, a very recent cutting-edge SSL framework, was proposed in \cite{berthelot2021adamatch} to tackle the issue of SSL performance degradation when the class distribution differed between the labeled and unlabeled sets. This problem could ideally be solved by having the same class distribution between the pseudo-labels of the unlabeled data and the actual unlabeled data. However, because the real class distribution of unlabeled data is often unknown, AdaMatch estimates it using the class distribution of labeled data. To do so, AdaMatch first acquires the class distribution of a model's predictions on weakly-augmented labeled data. Similarly, it obtains the predicted class distribution of the model's predictions on weakly-augmented unlabeled data. Following that, it calculates the ratio of labeled data class distribution to the \textit{expected} unlabeled data class distribution. This ratio is then used to modify the model's prediction on unlabeled data so that the adjusted class distribution of the unlabeled data would follow the labeled data class distribution. AdaMatch proposes a relative confidence threshold that is not only based on a user-defined value suggested in FixMatch \cite{sohn2020fixmatch}, but also depends on the model's confidence in its predictions on weakly labeled data. Although, the unsupervised term of AdaMatch is quite similar to the one used in FixMatch \cite{sohn2020fixmatch}, AdaMatch employs an additional supervised cross-entropy loss between the model's prediction on strongly-augmented labeled data and ground truth. AdaMatch outperformed other approaches, including FixMatch, on various computer vision datasets and achieved state-of-the-art results \cite{berthelot2021adamatch}.

\begin{table*}[ht]
    \centering
    \setlength
    \tabcolsep{10pt}
    \caption{The accuracy (in \%) of PARSE in comparison to other semi-supervised methods on the SEED dataset.}
    \begin{tabular}{l|llllll}
    \toprule
        Method                                  & 1 label               & 3 labels              & 5 labels              & 7 labels              & 10 labels             & 25 labels             \\ 
        \hline \hline
        $\Pi$-model \cite{samuli2017temporal}	&	60.25 $ $ \tiny(9.57)	&	67.87 \tiny(10.14)	&	72.43 \tiny(11.32)	&	74.94 \tiny(10.84)	&	76.35 \tiny(10.93)	&	77.87 \tiny(10.88)	\\	
        
        Temporal Ens. \cite{samuli2017temporal}	&	59.22 $ $ \tiny(9.02)	&	69.95 $ $ \tiny(9.07)	&	73.80 $ $ \tiny(9.78)	&	77.15 $ $ \tiny(9.57)	&	79.80 $ $ \tiny(9.53)	&	83.83 $ $ \tiny(8.73)	\\	
        
        Mean Teacher \cite{tarvainen2017mean}	&	53.97 $ $ \tiny(8.24)	&	62.75 $ $ \tiny(9.98)	&	66.42 $ $ \tiny(9.46)	&	69.90 \tiny(11.32)	&	71.48 $ $ \tiny(8.98)	&	77.09 $ $ \tiny(9.66)	\\	
        
        Conv. AutoEnc. \cite{tong2019caesnet}	&	71.39 \tiny(12.20)	&	80.03 \tiny(11.69)	&	82.86 \tiny(10.89)	&	84.74 $ $ \tiny(9.70)	&	85.46 $ $ \tiny(9.77)	&	87.34 $ $ \tiny(8.96)	\\	
        
        Pseudo-Label  \cite{lee2013pseudo}	&	68.02 \tiny(13.20)	&	78.11 \tiny(12.02)	&	79.57 \tiny(10.78)	&	82.21 \tiny(11.03)	&	84.11 $ $ \tiny(9.79)	&	85.32 $ $ \tiny(9.38)	\\	
        
        MixMatch \cite{berthelot2019mixmatch}	&	68.97 \tiny(13.93)	&	80.89 \tiny(12.80)	&	\underline{83.94 \tiny(10.30)}	&	85.46 $ $ \tiny(9.64)	&	85.84 $ $ \tiny(9.24)	&	86.88 $ $ \tiny(8.78)	\\	
        
        FixMatch  \cite{sohn2020fixmatch}	&	66.36 \tiny(13.84)	&	76.26 \tiny(11.56)	&	79.04 \tiny(10.68)	&	81.79 \tiny(10.56)	&	83.14 $ $ \tiny(9.98)	&	84.44 $ $ \tiny(9.09)	\\	
        
        AdaMatch   \cite{berthelot2021adamatch}	&	\underline{74.03 \tiny(11.78)}	&	\underline{82.59 \tiny(10.26)}	&	83.62 \tiny(10.84)	&	\underline{85.84 $ $ \tiny(9.69)}	&	\underline{86.71 $ $ \tiny(9.09)}	&	\underline{88.02 $ $ \tiny(8.80)}	\\	
        \hline
        \textbf{PARSE (ours)}	&	\textbf{77.77 \tiny(12.05)}	&	\textbf{86.52 \tiny(10.02)}	&	\textbf{88.38 $ $ \tiny(9.09)}	&	\textbf{89.55 $ $ \tiny(8.62)}	&	\textbf{90.50 $ $ \tiny(7.74)}	&	\textbf{91.14 $ $ \tiny(7.52)}	\\	
    \bottomrule
    \end{tabular}
    \label{seed}
\end{table*}

\begin{table*}[ht]
    \centering
    \setlength
    \tabcolsep{10pt}
    \caption{The accuracy (in \%) of PARSE in comparison to other semi-supervised methods on SEED-IV dataset.}
    \begin{tabular}{l|llllll}
    \toprule
        Method                                  & 1 label               & 3 labels              & 5 labels              & 7 labels              & 10 labels             & 25 labels             \\
        \hline \hline
        $\Pi$-model \cite{samuli2017temporal}	&	49.93 \tiny(12.30)	&	54.65 \tiny(14.66)	&	57.65 \tiny(14.36)	&	58.79 \tiny(14.77)	&	60.14 \tiny(15.14)	&	61.92 \tiny(15.30)	\\	
        
        Temporal Ens. \cite{samuli2017temporal}	&	52.76 \tiny(13.15)	&	59.76 \tiny(14.48)	&	63.00 \tiny(14.35)	&	65.26 \tiny(14.07)	&	65.92 \tiny(13.71)	&	67.25 \tiny(13.63)	\\	
        
        Mean Teacher \cite{tarvainen2017mean}	&	47.03 \tiny(11.84)	&	51.56 \tiny(12.35)	&	55.05 \tiny(13.27)	&	56.60 \tiny(12.91)	&	56.66 \tiny(11.78)	&	57.97 \tiny(13.00)	\\	
        
        Conv. AutoEnc. \cite{tong2019caesnet}	&	53.19 \tiny(18.58)	&	59.52 \tiny(18.13)	&	63.01 \tiny(16.74)	&	64.83 \tiny(15.75)	&	66.40 \tiny(17.26)	&	65.96 \tiny(16.62)	\\	
        
        Pseudo-Label  \cite{lee2013pseudo}	&	52.31 \tiny(17.93)	&	58.08 \tiny(16.76)	&	60.36 \tiny(17.92)	&	60.84 \tiny(17.59)	&	62.13 \tiny(19.03)	&	62.71 \tiny(18.36)	\\	
        
        MixMatch \cite{berthelot2019mixmatch}	&	56.08 \tiny(15.92)	&	65.03 \tiny(15.79)	&	\underline{69.42 \tiny(16.31)}	&	\textbf{70.92 \tiny(16.02)}	&	\textbf{72.31 \tiny(16.27)}	&	\textbf{73.20 \tiny(15.19)}	\\	
        
        FixMatch  \cite{sohn2020fixmatch}	&	53.37 \tiny(17.33)	&	63.57 \tiny(15.57)	&	63.43 \tiny(16.26)	&	64.62 \tiny(15.57)	&	66.50 \tiny(15.91)	&	68.54 \tiny(15.58)	\\	
        
        AdaMatch   \cite{berthelot2021adamatch}	&	\textbf{58.30 \tiny(15.95)}	&\underline{66.52 \tiny(16.58)}	&	69.12 \tiny(16.45)	&	68.11 \tiny(15.80)	&	69.31 \tiny(16.87)	&	71.43 \tiny(16.04)	\\	
        \hline
        \textbf{PARSE (ours)}	&	\underline{57.87 \tiny(18.09)}	&	\textbf{68.53 \tiny(16.34)}	&	\textbf{69.66 \tiny(15.96)}	&	\underline{70.77 \tiny(16.47)}	&	\underline{72.15 \tiny(16.08)}	&	\underline{72.32 \tiny(16.15)}	\\	
        
    \bottomrule
    \end{tabular}
    \label{seed-iv}
\end{table*}

\begin{table*}[ht]
    \centering
    \setlength
    \tabcolsep{10pt}
    \caption{The accuracy (in \%) of PARSE in comparison to other semi-supervised methods on SEED-V dataset.}
    \begin{tabular}{l|llllll}
    \toprule
        Method                                  & 1 label               & 3 labels              & 5 labels              & 7 labels              & 10 labels             & 25 labels             \\
        \hline \hline
        $\Pi$-model \cite{samuli2017temporal}	&	36.73 $ $ \tiny(8.17)	&	43.07 \tiny(10.47)	&	47.39 \tiny(11.04)	&	48.09 \tiny(11.32)	&	49.19 \tiny(11.58)	&	51.87 \tiny(12.80)	\\	

        Temporal Ens. \cite{samuli2017temporal}	&	\underline{39.09 $ $ \tiny(8.63)}	&	48.68 \tiny(10.52)	&	54.70 \tiny(12.19)	&	56.88 \tiny(12.75)	&	59.45 \tiny(12.68)	&	62.58 \tiny(12.37)	\\	
        
        Mean Teacher \cite{tarvainen2017mean}	&	36.13 $ $ \tiny(7.84)	&	42.60 $ $ \tiny(8.99)	&	46.34 \tiny(10.27)	&	45.91 \tiny(10.93)	&	48.38 \tiny(10.94)	&	50.79 \tiny(11.16)	\\	
        
        Conv. AutoEnc. \cite{tong2019caesnet}	&	36.08 \tiny(12.12)	&	50.30 \tiny(14.87)	&	58.38 \tiny(14.67)	&	60.94 \tiny(15.12)	&	65.47 \tiny(15.36)	&	67.63 \tiny(15.48)	\\	
        
        Pseudo-Label  \cite{lee2013pseudo}	&	32.59 \tiny(12.13)	&	49.04 \tiny(14.33)	&	51.75 \tiny(16.29)	&	56.13 \tiny(15.99)	&	59.62 \tiny(16.20)	&	61.70 \tiny(16.24)	\\	
        
        MixMatch \cite{berthelot2019mixmatch}	&	34.82 $ $ \tiny(9.14)	&	\underline{54.76 \tiny(12.50)}	&	\textbf{63.04 \tiny(13.09)}	&	\textbf{67.96 \tiny(13.17)}	&	\textbf{70.40 \tiny(13.00)}	&	\textbf{74.27 \tiny(13.02)}	\\	
        
        FixMatch  \cite{sohn2020fixmatch}	&	38.62 \tiny(10.74)	&	52.28 \tiny(12.40)	&	59.28 \tiny(13.00)	&	62.11 \tiny(13.92)	&	65.47 \tiny(13.18)	&	67.92 \tiny(13.08)	\\	
        
        AdaMatch   \cite{berthelot2021adamatch}	&	37.46 \tiny(11.29)	&	52.63 \tiny(13.80)	&	59.67 \tiny(14.23)	&	62.77 \tiny(15.34)	&	65.08 \tiny(15.08)	&	68.03 \tiny(15.97)	\\	
        \hline
        \textbf{PARSE (ours)}	&	\textbf{39.42 $ $ \tiny(9.79)}	&	\textbf{55.66 \tiny(12.72)}	&	\underline{62.40 \tiny(13.19)}	&	\underline{65.76 \tiny(13.04)}	&	\underline{68.73 \tiny(14.00)}	&	\underline{71.50 \tiny(14.05)}	\\	

    \bottomrule
    \end{tabular}
    \label{seed-v}
\end{table*}

\begin{table*}[ht]
    \centering
    \setlength
    \tabcolsep{10pt}
    \caption{The F1-score (in \%) of PARSE in comparison to other semi-supervised methods on AMIGOS dataset (valence and arousal).}
    \begin{tabular}{l|llllll}
    \toprule
        & \multicolumn{6}{c}{Valence} \\
        \hline 
        
        Method                                  & 1 label               & 3 labels              & 5 labels              & 7 labels              & 10 labels             & 25 labels             \\
        \hline\hline
        $\Pi$-model \cite{samuli2017temporal}	&	42.01 $ $ \tiny(9.32)	&	46.14 $ $ \tiny(7.14)	&	45.58 $ $ \tiny(7.03)	&	46.06 $ $ \tiny(7.81)	&	46.99 $ $ \tiny(7.28)	&	46.31 $ $ \tiny(6.57)	\\	
        
        Temporal Ens. \cite{samuli2017temporal}	&	45.14 $ $ \tiny(8.11)	&	49.24 $ $ \tiny(7.93)	&	49.62 $ $ \tiny(7.01)	&	51.27 $ $ \tiny(6.59)	&	52.10 $ $ \tiny(6.70)	&	51.62 $ $ \tiny(6.28)	\\	
        
        Mean Teacher \cite{tarvainen2017mean}	&	\textbf{52.54 $ $ \tiny(7.72)}	&	\underline{53.14 $ $ \tiny(7.60)}	&	\underline{54.05 $ $ \tiny(6.81)}	&	53.84 $ $ \tiny(7.55)	&	54.46 $ $ \tiny(6.51)	&	54.75 $ $ \tiny(6.43)	\\	
        
        Conv. AutoEnc. \cite{tong2019caesnet}	&	39.69 \tiny(10.81)	&	45.92 \tiny(10.14)	&	48.20 $ $ \tiny(9.82)	&	48.73 $ $ \tiny(9.68)	&	49.53 \tiny(10.55)	&	53.02 $ $ \tiny(9.89)	\\	
        
        Pseudo-Label  \cite{lee2013pseudo}	&	40.69 $ $ \tiny(8.18)	&	44.34 $ $ \tiny(9.35)	&	45.16 $ $ \tiny(9.26)	&	46.83 $ $ \tiny(9.91)	&	46.03 $ $ \tiny(9.15)	&	47.68 $ $ \tiny(9.86)	\\	
        
        MixMatch \cite{berthelot2019mixmatch}	&	45.81 \tiny(20.08)	&	50.56 \tiny(16.90)	&	\textbf{54.32 \tiny(13.81)}	&	\underline{53.86 \tiny(14.03)}	&	\underline{55.58 \tiny(11.22)}	&	55.94 \tiny(10.45)	\\	
        
        FixMatch  \cite{sohn2020fixmatch}	&	40.01 $ $ \tiny(8.75)	&	45.48 \tiny(11.34)	&	46.17 \tiny(10.15)	&	47.96 \tiny(10.82)	&	48.41 \tiny(10.48)	&	50.10 $ $ \tiny(9.69)	\\	
        
        AdaMatch   \cite{berthelot2021adamatch}	&	43.56 \tiny(11.10)	&	49.67 \tiny(10.83)	&	50.43 \tiny(10.53)	&	50.90 \tiny(10.50)	&	53.53 \tiny(10.89)	&	\underline{56.53 \tiny(10.65)}	\\	
        \hline
        
        \textbf{PARSE (ours)}	&	\underline{47.53 $ $ \tiny(9.46)}	&	\textbf{55.02 \tiny(12.68)}	&	53.43 \tiny(10.91)	&	\textbf{56.18 \tiny(10.15)}	&	\textbf{55.92 $ $ \tiny(9.72)}	&	\textbf{58.77 \tiny(10.80)}	\\

        \toprule
       
        & \multicolumn{6}{c}{Arousal} \\
        \hline
        
        Method                                  & 1 label               & 3 labels              & 5 labels              & 7 labels              & 10 labels             & 25 labels             \\
        \hline\hline
        $\Pi$-model \cite{samuli2017temporal}	&	43.12 \tiny(10.17)	&	44.66 $ $ \tiny(6.58)	&	45.12 $ $ \tiny(7.07)	&	46.31 $ $ \tiny(6.77)	&	46.87 $ $ \tiny(7.08)	&	47.36 $ $ \tiny(6.47)	\\	
        
        Temporal Ens. \cite{samuli2017temporal}	&	46.49 $ $ \tiny(6.82)	&	49.37 $ $ \tiny(6.23)	&	50.52 $ $ \tiny(6.81)	&	50.98 $ $ \tiny(5.70)	&	51.10 $ $ \tiny(5.84)	&	50.78 $ $ \tiny(5.44)	\\	
        
        Mean Teacher \cite{tarvainen2017mean}	&	\textbf{50.95 $ $ \tiny(7.24)}	&	\underline{52.85 $ $ \tiny(7.03)}	&	\underline{53.38 $ $ \tiny(6.71)}	&	\underline{54.05 $ $ \tiny(6.09)}	&	54.27 $ $ \tiny(5.23)	&	54.61 $ $ \tiny(5.68)	\\	
        
        Conv. AutoEnc. \cite{tong2019caesnet}	&	40.78 \tiny(10.05)	&	46.79 $ $ \tiny(9.17)	&	47.45 $ $ \tiny(7.39)	&	48.58 $ $ \tiny(8.45)	&	49.47 $ $ \tiny(9.03)	&	50.68 $ $ \tiny(9.08)	\\	
        
        Pseudo-Label  \cite{lee2013pseudo}	&	43.27 $ $ \tiny(7.62)	&	45.19 $ $ \tiny(7.85)	&	45.73 $ $ \tiny(7.82)	&	46.48 $ $ \tiny(7.26)	&	47.03 $ $ \tiny(7.35)	&	48.32 $ $ \tiny(8.94)	\\	
        
        MixMatch \cite{berthelot2019mixmatch}	&	\underline{48.76 \tiny(22.42)}	&	\textbf{53.73 \tiny(19.16)}	&	\textbf{56.70 \tiny(16.84)}	&	\textbf{57.13 \tiny(16.18)}	&	\textbf{59.11 \tiny(12.97)}	&	\textbf{59.58 \tiny(12.29)}	\\	
        
        FixMatch  \cite{sohn2020fixmatch}	&	40.49 \tiny(10.61)	&	45.07 $ $ \tiny(8.57)	&	47.68 $ $ \tiny(7.20)	&	48.08 $ $ \tiny(7.48)	&	48.90 $ $ \tiny(8.22)	&	50.34 $ $ \tiny(8.66)	\\	
        
        AdaMatch   \cite{berthelot2021adamatch}	&	40.09 \tiny(10.24)	&	49.28 \tiny(11.35)	&	50.92 \tiny(10.76)	&	52.20 \tiny(10.30)	&	53.44 \tiny(10.96)	&	\underline{58.50 \tiny(10.04)}	\\	
        \hline
        
        \textbf{PARSE (ours)}	&	48.46 \tiny(10.47)	&	52.63 \tiny(10.21)	&	52.51 $ $ \tiny(9.44)	&	53.77 $ $ \tiny(9.91)	&	\underline{55.21 $ $ \tiny(8.85)}	&	57.76 $ $ \tiny(8.66)	\\

        \bottomrule

    \end{tabular}
    \label{amigos}
\end{table*}

\begin{figure*}
    \begin{center}
    \includegraphics[width=1.0\textwidth]{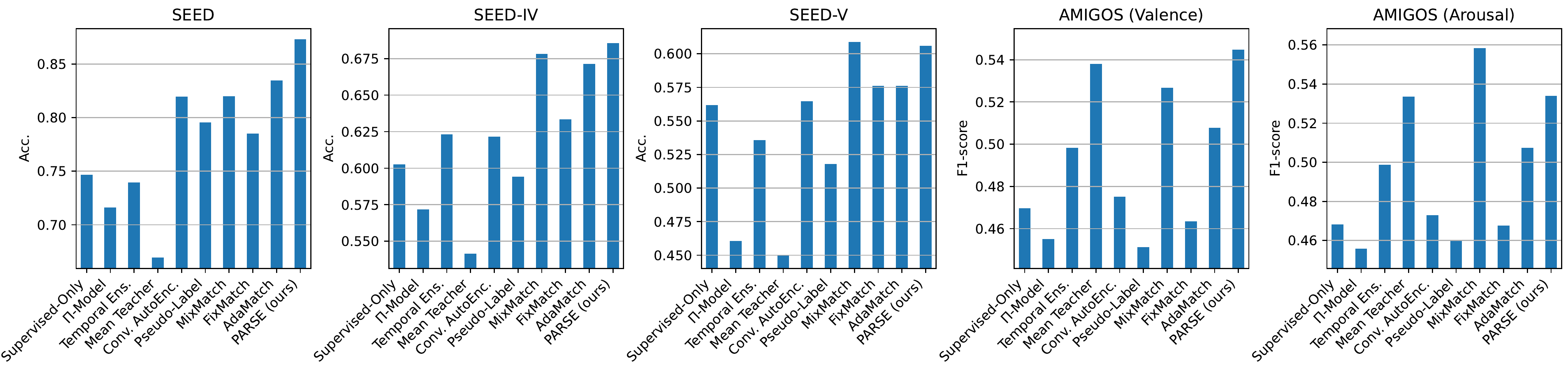} 
    \caption{Average performance of PARSE in comparison to other SSL methods as well as supervised learning method for SEED, SEED-IV, SEED-V, and AMIGOS (valence and arousal), across all $6$ few-labeled scenarios ($m \in \{1,3,5,7,10,25\}$).}
    \label{comparison}
    \end{center}
\end{figure*}

\section{Results and Discussions} \label{sec: results and discussion}

\subsection{Experimental Results} \label{sec: experimental results}
We compare our proposed framework to other existing methods when only few training samples per class are labeled ($m \in \{1, 3, 5, 7, 10, 25\}$). This experimental protocol is in accordance to recent state-of-the-art studies on SSL \cite{berthelot2019mixmatch,sohn2020fixmatch,berthelot2021adamatch}. Each method is evaluated \textbf{five} times for all $6$ few-labeled scenarios, each time with a different random seed for the selection of $D_l$. Tables \ref{seed}, \ref{seed-iv}, \ref{seed-v}, and \ref{amigos} show the average and standard deviation of the results for SEED, SEED-IV, SEED-V, and AMIGOS (valence and arousal) across $5$ different random seeds, respectively. Following, we provide the detailed results and discussions on each dataset.

\textbf{SEED.} As shown in Table \ref{seed}, classical SSL methods generally produce inferior performance in all $6$ designated few-labeled scenarios, except that convolutional autoencoder consistently outperforms other classical approaches and slightly outperforms FixMatch. In all of the few-labeled scenarios, AdaMatch consistently outperforms FixMatch and all classical SSL approaches by achieving the second-best result (shown with underline). Moreover, AdaMatch outperforms MixMatch only with the exception of $5$ labeled samples-per-class scenario. In the meantime, the convolutional autoencoder and MixMatch achieve the third-best results for $m\in \{1,25\}$ and $m\in \{3,7,10\}$, respectively. Our proposed method, PARSE, consistently achieves the best (shown in bold) performance, outperforming the second-best methods by more than $3.0\%$ across all the few-labeled scenarios, demonstrating its superiority in the absence of sufficient labeled samples.

\textbf{SEED-IV.} As shown in Table \ref{seed-iv}, traditional SSL methods perform worse than other more recent solutions. Temporal ensembling and convolutional autoencoder have comparable performances and produce the best results among classical approaches. MixMatch, FixMatch, and AdaMatch nearly outperform all of the classical methods across all $6$ few-labeled scenarios. In particular, when \textbf{only one} sample per class is labeled, PARSE achieves the second-best result, approaching the best method obtained by AdaMatch with only $0.43\%$ difference. Our PARSE achieves the best results when more labeled samples are given ($m \in \{3, 5\}$). Especially in the case of $3$ labeled samples per class, our method outperforms the second-best result (AdaMatch) by $2\%$. When even more labeled samples are provided, PARSE approaches the highest performance with a very small difference ($0.15\%$, $0.16\%$ and $0.88\%$ for $m \in \{7,10,25\}$).

\textbf{SEED-V.} As shown in Table \ref{seed-v}, the classical methods are generally outperformed by other recent techniques, except that temporal ensembling achieves the second-best result when $m=1$, and convolutional autoencoder closely approaches the third-best result in case of $m=10$. Furthermore, FixMatch and AdaMatch have similar performance across all the few-labeled scenarios. Our proposed architecture achieves the best results when $1$ and $3$ labeled samples per class are provided, respectively. In cases with more than $3$ labeled samples, MixMatch obtains the best results despite its poor performance in the barely supervised scenario ($m=1$) while PARSE consistently obtains the second-best results.

\begin{figure*}
    \begin{center}
    \includegraphics[width=1.0\textwidth]{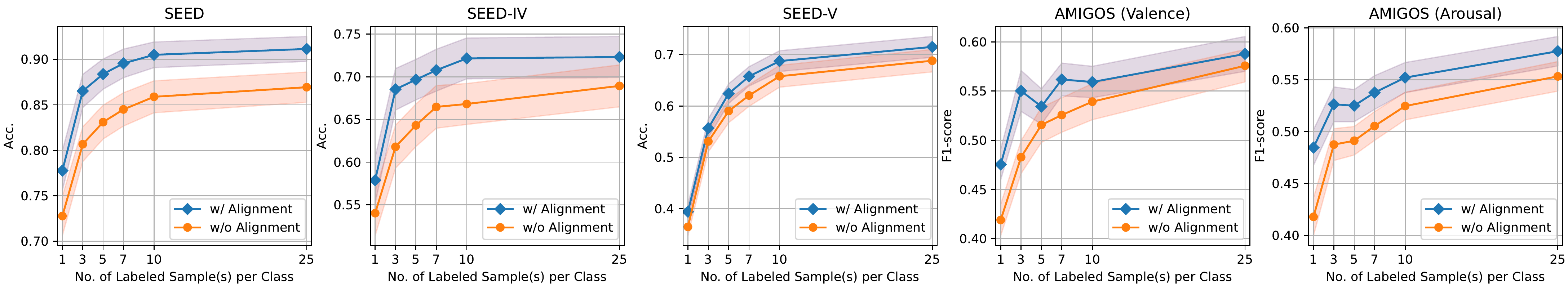} 
    \caption{The performance of PARSE with and without representation alignment, across varying number of labeled samples per class ($m \in \{1,3,5,7,10,25\}$) in SEED, SEED-IV, SEED-V and AMIGOS (valence and arousal).}
    \label{ablation}
    \end{center}
\end{figure*}

\begin{figure*}
    \begin{center}
    \includegraphics[width=1\textwidth]{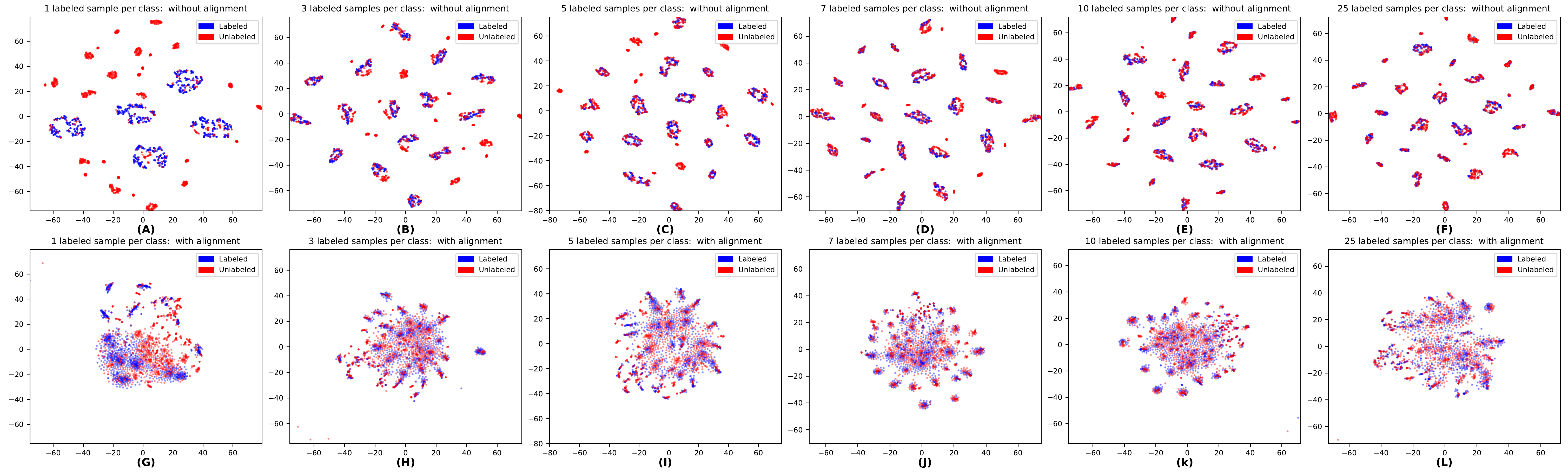} 
    \caption{Visual comparison, using t-SNE, between learned EEG embeddings without representation alignment ($1^{st}$ row) and with representation alignment ($2^{nd}$ row) when a varying number of labeled samples per class are used ($m \in \{1,3,5,7,10,25\}$). Here, the size of labeled and unlabeled sets are the same due to the replication process applied to the labeled set to increase their size (see Section \ref{sec: our approach}). We observe stronger overlaps between the labeled and unlabeled samples when representation alignment is applied in PARSE, which will eventually result in more confident guessed labels for the unlabeled samples.}
    \label{tsne}
    \end{center}
\end{figure*}

\textbf{AMIGOS (valence).} As shown in Table \ref{amigos} (Valence), among the classical approaches, mean teacher method consistently achieves the highest performance, followed by temporal ensembling. Mean teacher also outperforms all other approaches in the barely supervised scenario, followed by our proposed method. MixMatch achieves the best result when $m=5$ and the second-best ones when more samples are labeled ($m \in \{7, 10\}$). Our PARSE achieves the best performance when $m \in \{3,7,10,25\}$ and approaches the best result at $m=5$ with less than $1\%$ difference.

\textbf{AMIGOS (arousal).} As displayed in Table \ref{amigos} (Arousal), similar to valence, mean teacher consistently achieves the highest results, followed by temporal ensembling among all of the classical SSL methods. Especially, mean teacher achieves the best result in the barely supervised scenario and obtains the second-best results among all other methods in cases of more labeled samples ($m \in\{3,5,7\}$). MixMatch achieves the second-best result in the barely supervised setting and achieves the best results in the remaining few-labeled scenarios. It can be observed that for estimating arousal in the AMIGOS, our method does not perform the best in any of the cases, yet consistently obtains competitive performances all around.

\subsection{Comparison} \label{sec: comparison}
In Figure \ref{comparison}, we compare the average performance (across $m \in \{1,3,5,7,10,25\}$) for all the SSL methods as well as a baseline supervised-only method. In this experiment, we evaluate a supervised method including an encoder and classifier with the same architecture used in the SSL methods and trained only on labeled data (without any unlabeled data). 

Among the classical SSL methods with consistency regularization technique, temporal ensembling outperforms the $\Pi$-model on all of the datasets. This is mainly due to the more reliable outputs being generated by temporal ensembling methods (as described in Section \ref{sec: temporal ensembling}). Meanwhile, it is interesting to note that mean teacher method performs worse than $\Pi$-model and temporal ensembling on SEED-series datasets with balanced class distributions while outperforming both on AMIGOS with highly unbalanced label distributions. We also observed that the pseudo-labeling method has moderate performance among all classical SSL methods on SEED-series datasets but performs poorly on AMIGOS (valence and arousal).
The large difference in label distribution between labeled and unlabeled data in AMIGOS is likely to causes the network (which was initially trained on labeled data) to generate pseudo-labels with the same class distribution as in the labeled set. Similarly, FixMatch performs well on the datasets with balanced class distributions but poorly on those with unbalanced ones. This is mainly because FixMatch ignores the class distribution mismatch between labeled and unlabeled sets (as described in Section \ref{sec: fixmatch}). AdaMatch, unlike FixMatch, takes into account class distribution alignment (as described in Section \ref{sec: adamatch}), delivering good results across all the datasets. Compared to FixMatch and AdaMatch, MixMatch does not consider the distribution alignment but encourages convex combinations of labeled and unlabeled samples (as described in Section \ref{sec: mixmatch}), leading to more consistent and higher performances across datasets.

Compared to all these cutting-edge semi-supervised solutions, our proposed PARSE considers both the confidence of pseudo-labels and the distribution alignment of labeled and unlabeled data. For the confidence of pseudo-labels, we use a warm-up function to slowly increase the unsupervised term's weight so that low-confidence pseudo-labels are rejected. For pairwise distribution alignment, our proposed module reduces the distribution mismatch between labeled and unlabeled data. As shown in Figure \ref{comparison}, our novel solution enables the model to perform better than MixMatch on SEED, SEED-IV, and AMIGOS (valance), as well as FixMatch and AdaMatch on all the datasets. Although MixMatch performs better than PARSE on AMIGOS (arousal), our model is much more stable, as evidenced by the considerably larger standard deviation of MixMatch on this dataset (see Table \ref{amigos} (Arousal)).

\begin{figure}
    \begin{center}
    \includegraphics[width=0.95\columnwidth]{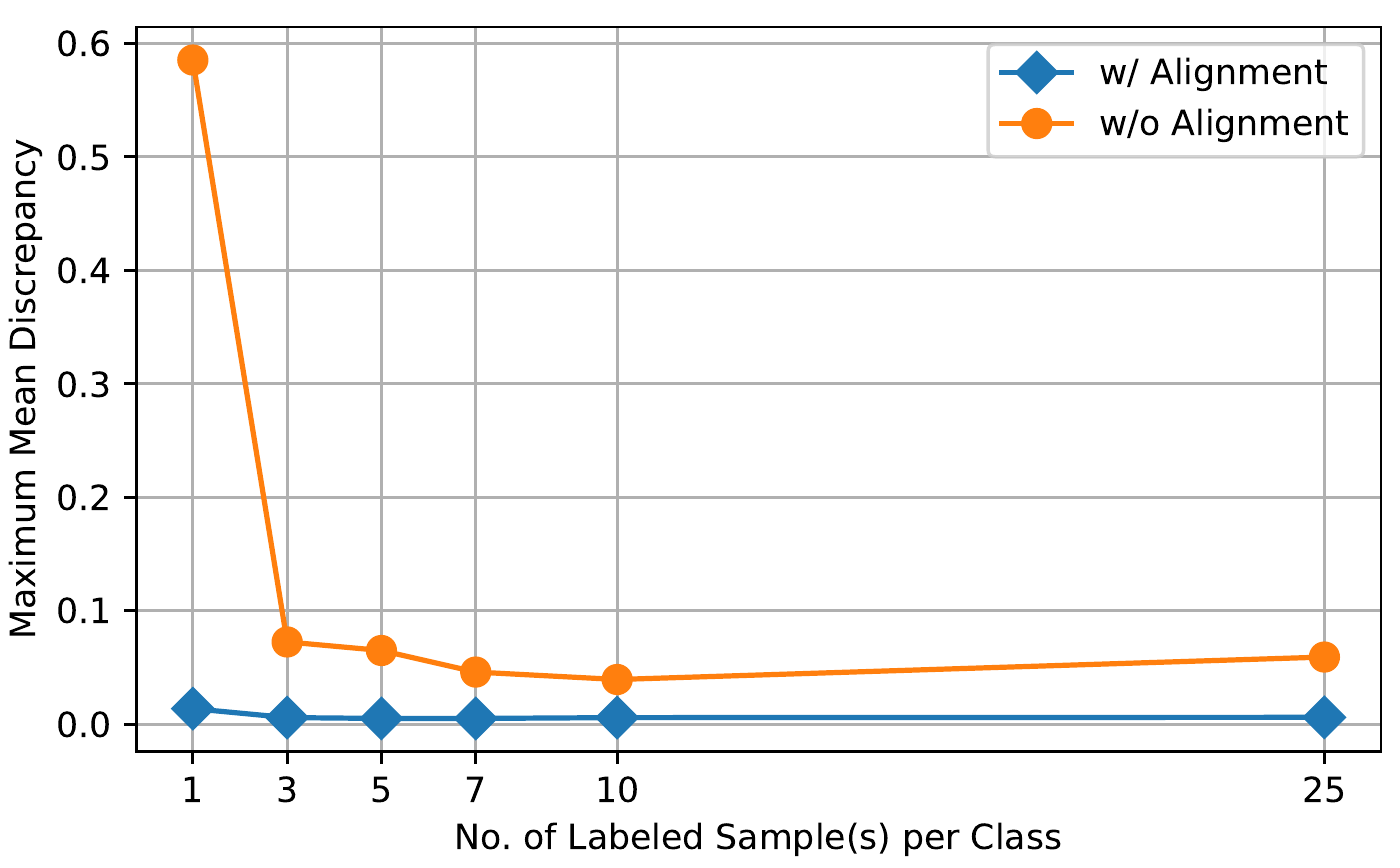} 
    \caption{Quantitative evaluation of the distances between labeled and unlabeled representations with and without alignment when a varying number of labeled samples ($m \in {1,3,5,7,10,25}$) are provided. MMD is used to calculate the distances.}
    \label{mmd}
    \end{center}
\end{figure}

\begin{figure*}
    \begin{center}
    \includegraphics[width=1.0\textwidth]{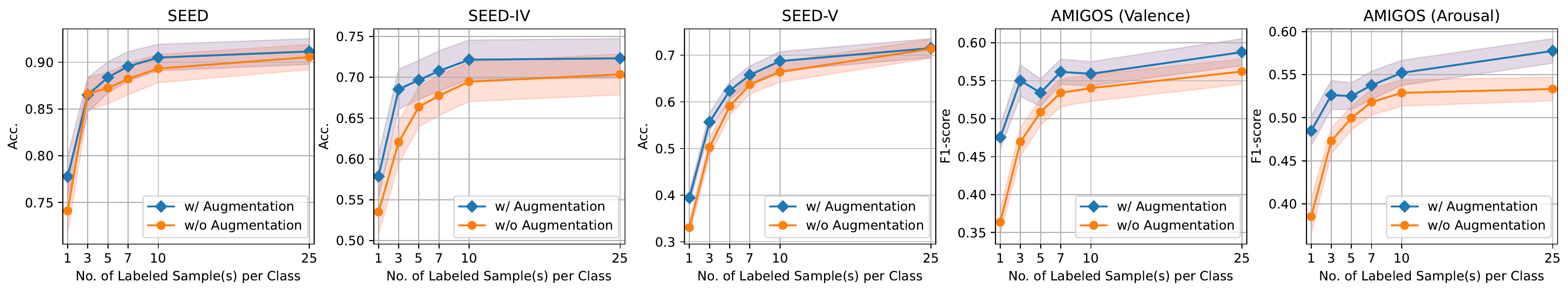} 
    \caption{The performance of PARSE with and without data augmentation by additive Gaussian noise, across varying number of labeled samples per class ($m \in \{1,3,5,7,10,25\}$) in SEED, SEED-IV, SEED-V, and AMIGOS (valence and arousal).}
    \label{aug}
    \end{center}
\end{figure*}

\begin{figure}
    \begin{center}
    \includegraphics[width=0.40\textwidth]{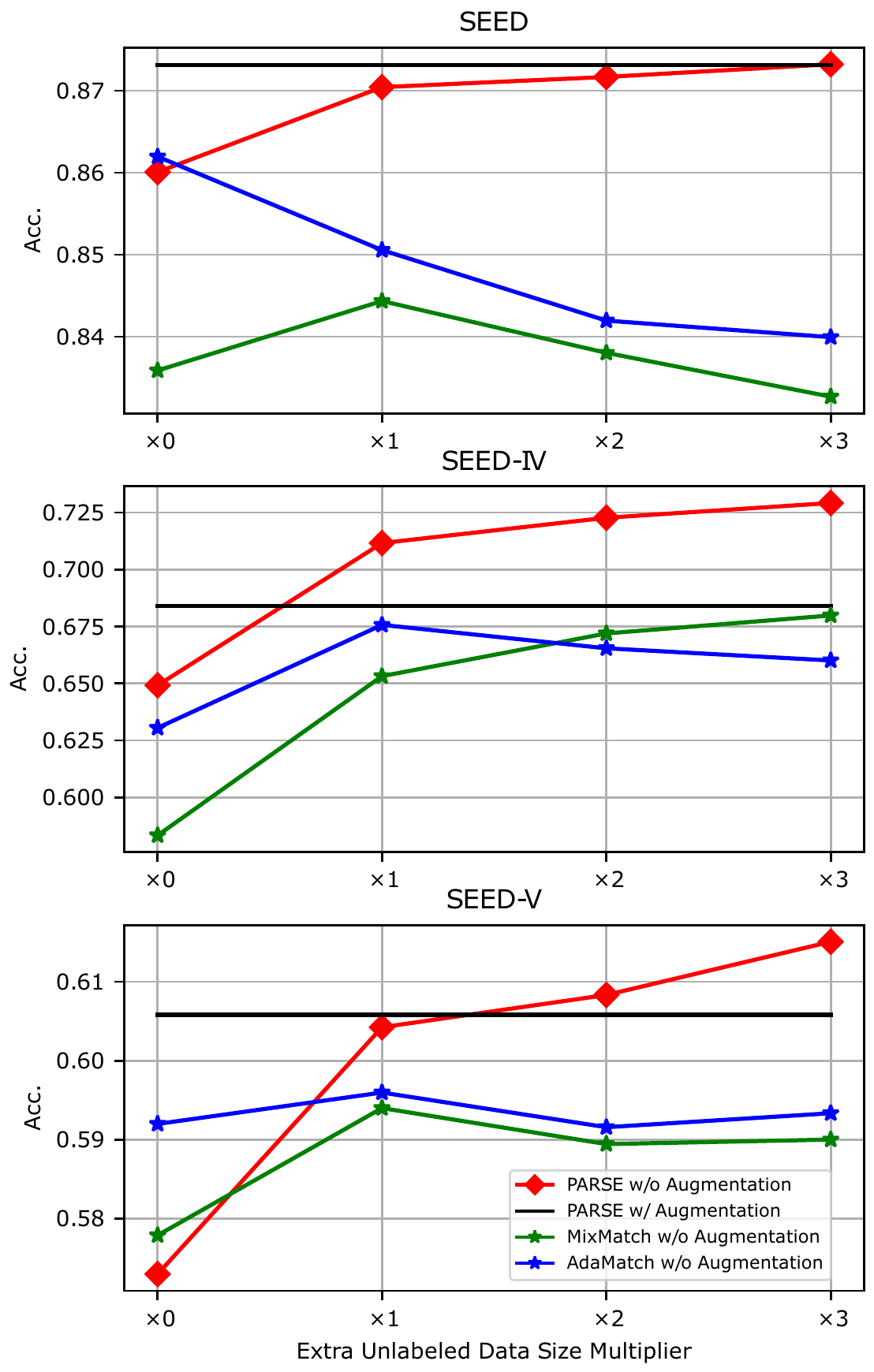} 
    \caption{The average performance of PARSE in comparison to other top-performing SSL methods (MixMatch and AdaMatch) when data augmentation is removed, across varying numbers of extra unlabeled data (whose size are $\times$1, $\times$2, and $\times$3 larger than the original unlabeled dataset size) in SEED-series datasets. The results of all the aforementioned approaches without extra unlabeled data ($\times$0), and PARSE with augmentation but without extra unlabeled data (black line) are provided as references.}
    \label{extra_unlabeled_size}
    \end{center}
\end{figure}

\begin{figure*}
    \begin{center}
    \includegraphics[width=1.0\textwidth]{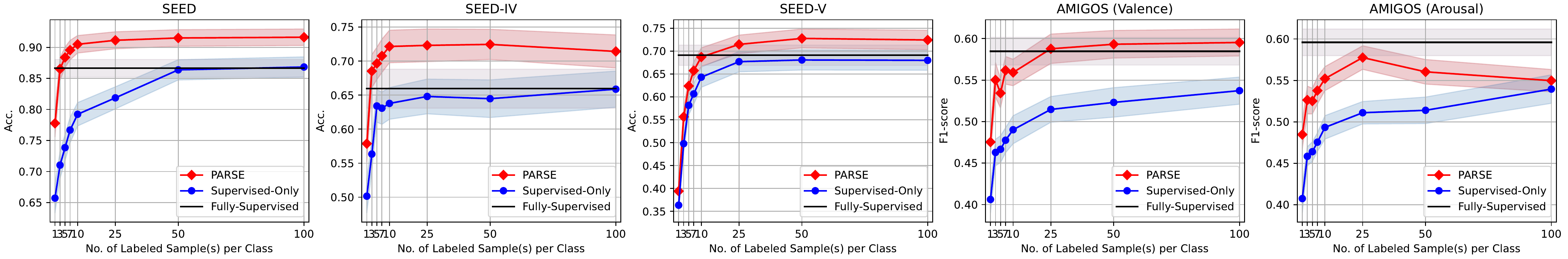} 
    \caption{The performance of PARSE in comparison to supervised-only method, across varying number of labeled samples per class ($m \in \{1,3,5,7,10,25,50, 100\}$) in SEED, SEED-IV, SEED-V and AMIGOS (valence and arousal). Fully-supervised method results are provided as references.}
    \label{comparison_fsl}
    \end{center}
\end{figure*}

As shown in Figure \ref{comparison}, on all the datasets, our proposed method outperforms the supervised-only model by considerable margins. It should be pointed out that unlike PARSE, not all SSL methods consistently outperform the supervised-only approaches. Moreover, a general observation is that the performance of most SSL methods is quite dataset-dependant, and while our performance also varies across different datasets, PARSE achieves more consistent results. This demonstrates the superiority of our method when faced with a scarcity of labeled samples across different datasets.

\subsection{Ablation and Analysis} \label{sec: ablation and analysis}
We perform an ablation study to investigate the impact of the pairwise representation alignment step in our method. Figure \ref{ablation} shows the performance (mean and standard error) of our proposed method with and without pairwise representation alignment for SEED, SEED-IV, SEED-V, and AMIGOS (valence and arousal). Over all of the datasets and across all the $6$ few-labeled scenarios, we find our model performing consistently better with pairwise representation alignment than without it. Specifically, when the class distribution of a dataset is balanced, the alignment improves PARSE, for example, $5.0\%$, $4.8\%$, and $3.0\%$ in SEED, SEED-IV, and SEED-V, respectively. Moreover, when the class distribution is imbalanced, the representation alignment boosts the performance by around $3.5\%$ and $3.7\%$ in AMIGOS (valence) and AMIGOS (arousal), respectively.

Following, we explore the learned EEG embeddings for both labeled and unlabeled data to better understand the impact of the representation alignment step in our model. To this end, we use t-distributed Stochastic Neighbor Embedding (t-SNE) to visualize the learned EEG representations. Figure \ref{tsne} shows a comparison between the representations with and without alignment for the different numbers of labeled samples in SEED-V (containing $5$ emotion classes) as examples. We observe in Figures \ref{tsne} (A) and (G) that when only $1$ labeled sample per class is available, the learned embeddings of the unlabeled data are far apart from the labeled data in the absence of pairwise representation alignment. However, we observe that when alignment is performed, these embeddings (unlabeled and labeled) are brought closer to each other. This is a desired property as it means that the output class information for the labeled samples can confidently be used to guess the labels for the unlabeled samples. We observe a similar, but less significant improvement, when more labeled samples are available ($m \in \{3,5,7,10,25\}$), (see Figures \ref{tsne} (B) through (F) vs. (H) through (L)). To provide quantitative evidence for this experiment, we employ the Maximum Mean Discrepancy (MMD) to measure the distance between the learned EEG representation of labeled and unlabeled sets. MMD values with and without pairwise representation alignment are compared across all $6$ few-labeled scenarios. As depicted in Figure \ref{mmd}, MMD is $0.0138$ and $0.5853$ with and without alignment when only $1$ labeled sample per class is available, highlighting the importance of representation alignment in the scarcely supervised situation. When more labeled samples are provided, however, MMD without alignment falls to $0.0726$ at $m=3$ but remains nearly constant thereafter (e.g., MMD of $0.0650$ at $m=5$). In all the few-labeled scenarios except for the barely supervised setting ($m > 1$), MMD values with alignment are consistently around $10 \times$ lower than MMD values without it. Our proposed method with pairwise representation alignment brings the labeled and unlabeled data closer to each other, particularly in the barely supervised scenario, resulting in consistent and improved performance. 

Next, we study the effect of data augmentation in our proposed framework. Figure \ref{aug} shows the performance (mean and standard error) of our proposed method with and without data augmentation by additive noise for SEED, SEED-IV, SEED-V, and AMIGOS (valence and arousal). We find that data augmentation improves the performance of our model in all the five datasets. In-line with previous research on semi-supervised learning \cite{samuli2017temporal,tarvainen2017mean,berthelot2019mixmatch,sohn2020fixmatch,berthelot2021adamatch}, this improvement is likely due to two reasons. First, data augmentation is a necessary component of reliable pseudo-label generation of unlabeled data (Eq. \ref{average_prediction}) \cite{berthelot2019mixmatch}. Second, we use model prediction of strongly augmented unlabeled data rather than that of the original unlabeled data for unsupervised consistency loss (as shown in Eq. \ref{l_u}) to prevent the loss from being too easily minimized, and the model from overfitting \cite{sohn2020fixmatch}. Interestingly, we also find that data augmentation has the most impact in the barely-supervised scenario, improving the performance by $3.7\%$, $6.4\%$, $11.2\%$, $9.9\%$ in SEED, SEED-V, AMIGOS (valence and arousal), respectively, while reaching the second-highest impact in SEED-IV with an improvement of $4.4\%$, among all $6$ few-labeled scenarios ($m \in \{1,3,5,7,10,25\}$).

Furthermore, we investigate the impact of using more \textit{real EEG data} instead of data augmentation in our model. To do so, we select additional unlabeled data from other subjects within the same dataset for each subject-dependent experiment in the SEED-series (SEED, SEED-IV, and SEED-V) datasets. Specifically, we choose the size of extra unlabeled data to be $\times$1, $\times$2, and $\times$3 larger than the original unlabeled dataset. Figure \ref{extra_unlabeled_size} shows the average performance of PARSE and two top-performing SSL methods (MixMatch and AdaMatch), across all $6$ few-labeled scenarios, over different amounts of extra real unlabeled data. In all the datasets, PARSE consistently outperforms MixMatch and AdaMatch by considerable margins. We also find that unlike MixMatch and AdaMatch, performance of PARSE without augmentation consistently increases when the size of extra unlabeled data is larger. Furthermore, with the help of enough extra unlabeled data, PARSE without augmentation can approach or even outperform our original version of PARSE (with augmentation). As a result, although the data augmentation strategy helps to improve PARSE's performance (see Figure \ref{aug}), we can further improve the performance by replacing data augmentation with more real unlabeled data (see Figure \ref{extra_unlabeled_size}), validating the effectiveness of our proposed semi-supervised learning framework.

\begin{figure}
    \begin{center}
    \includegraphics[width=0.95\columnwidth]{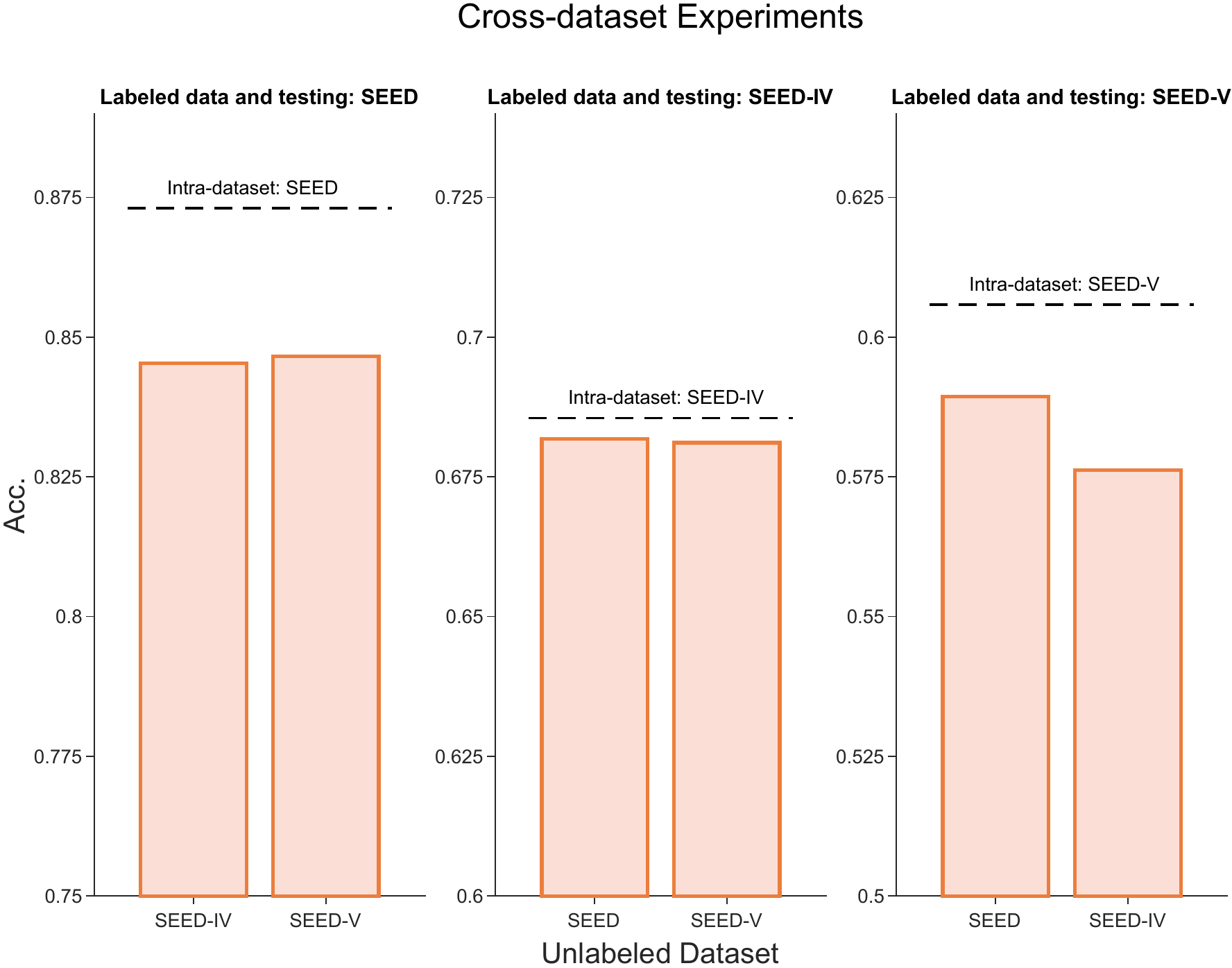} 
    \caption{Average performance of PARSE in cross-dataset experiments, across all $6$ few-labeled scenarios ($m \in \{1,3,5,7,10,25\}$).}
    \label{extra_unlabeled}
    \end{center}
\end{figure}

Next, we compare our proposed method against supervised training where the same number of labeled samples ($m \in \{1,3,5,7,10,25,50,100\}$) is used. We also explore a fully-supervised solution whereby training of the encoder and classifier is performed with the entire labeled training set. Our semi-supervised solution and its fully-supervised counterpart share the same training settings (in Section \ref{sec: implementation}) and the same architecture details.
We present the results in Figure \ref{comparison_fsl}, where we observe that PARSE outperforms the supervised-only method (using labeled samples only, without any unlabeled data) across all datasets and for varying numbers of $m$. We also look into the number of labeled samples needed for the performance of PARSE and the fully-supervised approach to become equal. We observe that in SEED, the two performances are almost equal at $3$ samples, whereas for SEED-IV the performances meet between $1$ and $3$ samples. It is observed that SEED-V requires more labeled samples for PARSE to meet and overtake the fully-supervised counterpart with approximately $10$ samples resulting in equal accuracies. Lastly, for the AMIGOS (valence) dataset, almost $25$ samples are required for PARSE and the fully-supervised to achieve equal performances.
The fact the semi-supervised model outperforms its fully-supervised counterpart (when we have large number of labeled samples), may be due to the semi-supervised model leveraging MixUp technique to widen the training data distribution through convex combinations between labeled and unlabeled data. This benefits the model predictions when test samples are out of the distribution of the original training set \cite{zhang2018mixup}.
However, interestingly for AMIGOS (arousal), our semi-supervised approach never achieves equal performance to, or overtakes, the fully-supervised counterpart. This is consistent with our findings in Table \ref{amigos} (Arousal) and Figure \ref{comparison} where despite achieving good results, our method performed slightly below MixMatch.

Furthermore, we examine the robustness of PARSE using unlabeled data from other datasets. We perform additional cross-dataset experiments using the SEED-series (SEED, SEED-IV, and SEED-V) datasets since they use the same EEG channels. Here, we perform three sub-experiments. First, we use the SEED dataset in labeled format, along with SEED-IV and SEED-V without their labels for training. We evaluate the model on SEED given the availability of labels. This is presented in Figure \ref{extra_unlabeled} (left). Next, we repeat the same experiment, this time selecting SEED-IV as the labeled dataset, while SEED and SEED-V are used as unlabeled datasets (evaluation is preformed on SEED-IV). The results for this experiment are depicted in Figure \ref{extra_unlabeled} (middle). Lastly, the experiment is repeated with SEED-V being selected as the labeled set, while SEED and SEED-IV are used without the labels for training, as evaluation is done on SEED-V. The results for this experiment are shown in Figure \ref{extra_unlabeled} (right). From these experiments, we conclude that PARSE performs very well and obtains competitive results to single-dataset tests, even when incorporating the set of \textit{unlabeled} data from other datasets.

\begin{figure}
    \begin{center}
    \includegraphics[width=1\columnwidth]{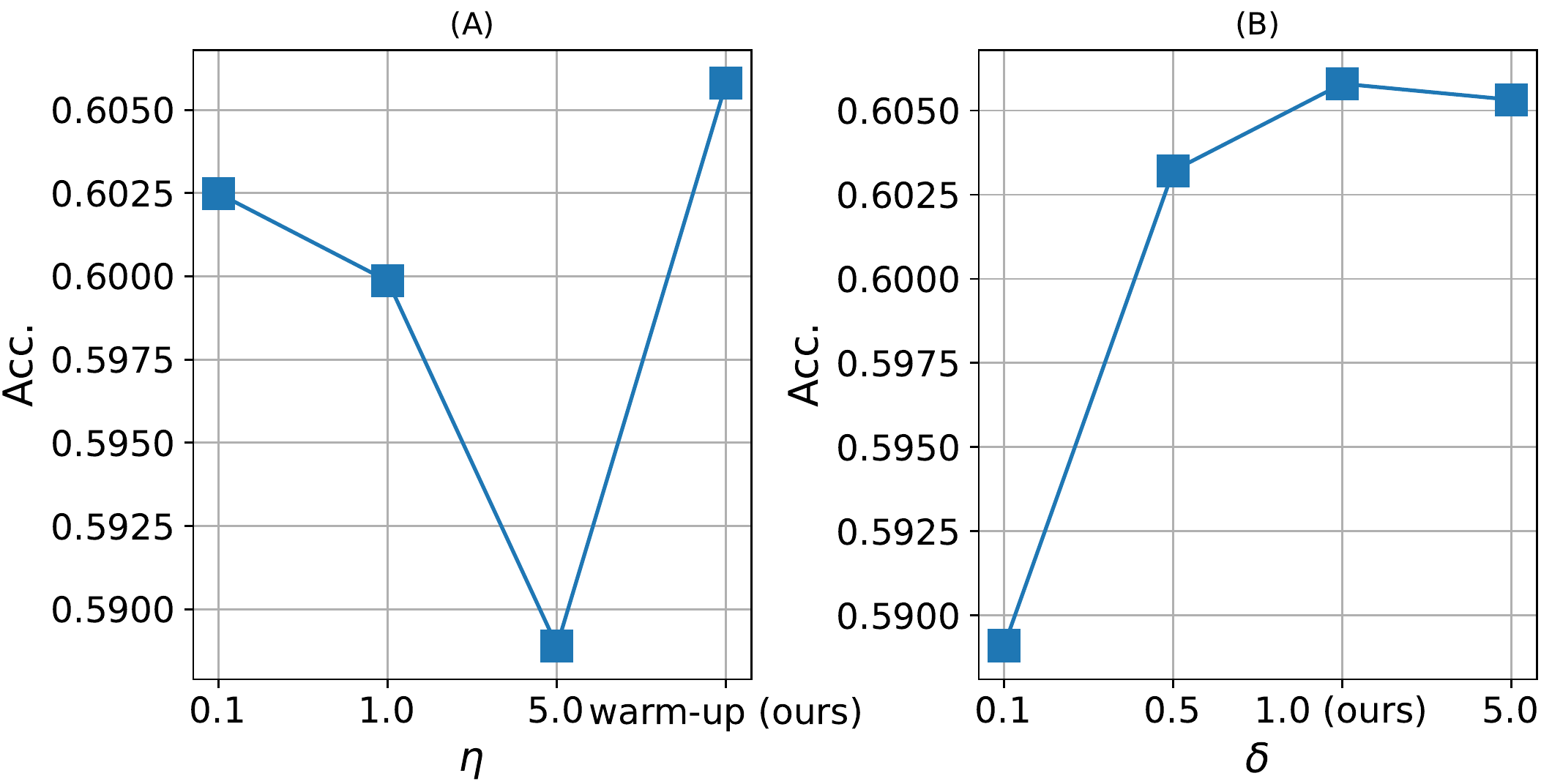} 
    \caption{The impact of different hyperparameters, $\eta$ (A) and $\delta$ (B), on the performance of PARSE averaged across all $6$ few-labeled scenarios.}
    \label{analysis}
    \end{center}
\end{figure}

Lastly, we perform further experiments to evaluate the effect of different hyper-parameter values on our model's performance. As described earlier (Eq. \ref{warm_up}), we use a warm-up function to gradually increase the weight $\eta$ applied to the unsupervised term. Here we evaluate the impact of this parameter in the final performance. Even though our warm-up function only selects $\eta$ values in the range of $0$ to $1$, manually selecting higher values is possible. Accordingly, we conduct our experiment by selecting $\eta = 0.1, 1.0, 5.0$, and compare them to ours. As illustrated in Figure \ref{analysis} (A), PARSE shows slight sensitivity to this parameter, and while small variations are observed by selecting different values, the warm-up function does provide the best results. Another parameter used in our model is $\delta$, which is the weight applied on the pairwise representation alignment loss term (Eq. \ref{total_loss}). To evaluate the impact of this parameter, we compare the performance of our method for different values of $\delta$ by setting it to $0.1, 0.5, 1.0, 5.0$. As shown in Figure \ref{analysis} (B), similar to $\eta$, $\delta$ does not have a significant impact on the performance of our model. Nonetheless, $\delta = 1.0$ does achieve the best performance, which we selected for our model.

\section{Conclusion and Future work} \label{sec: conclusion and future work}
In this research, we propose a novel semi-supervised method for EEG-based emotion recognition. Our model relies on data augmentation, label guessing, convex combinations of unlabeled and labeled sets, and pairwise representation alignment between the distributions of unlabeled and labeled data. We conduct extensive experiments against a number of other methods, with only $1,3,5,7,10$, and $25$ labeled samples per class, and evaluate the performance on four large publicly available datasets, SEED, SEED-IV, SEED-V, and AMIGOS (valence and arousal). Our proposed framework achieves average best results across all $6$ labeled scenarios in SEED, SEED-IV, and AMIGOS (valence), and closely approaches the best result with only a $0.3\%$ difference in SEED-V. Our method also reaches the second-best performance on AMIGOS (arousal). In addition, we show the impact of the pairwise representation alignment on our proposed method, with the varying number of labeled samples across all the datasets. We perform further analysis that shows our pairwise representation alignment considerably reduces the distance between labeled and unlabeled representations, especially in the barely supervised scenarios. The results also show that our framework consistently outperforms the supervised-only method, addressing the challenge of scarcity of labeled EEG data.

In-line with prior works, 
our analysis above shows that large amounts of labeled data are required for supervised EEG representation learning to achieve reliable performance. In practice, labeling each subject's EEG data is costly and time demanding. Moreover, the collected labels might be noisy and incomplete. Our proposed semi-supervised approach requires substantially less number of labeled samples to obtain a similar performance to fully-supervised models trained on large-scale labeled datasets. As a result, our proposed solution can alleviate the burden of future data annotation and labeling, either by self-reporting from the subjects or external professional annotators.

For future work, the challenges of potential label distribution mismatch and domain difference across datasets can be further explored and addressed. Moreover, the combination of partial label learning and semi-supervised learning can be investigated to alleviate the time-consuming process of assigning candidate labels to large amounts of training examples.

\section{Acknowledgement}
We would like to thank all of the reviewers for their suggestions and insights that helped to improve the quality of this research. 
\bibliographystyle{IEEEtran}
\bibliography{IEEEabrv,refs}

\end{document}